\documentclass[a4paper,fleqn]{cas-dc}

\usepackage[numbers]{natbib}
\usepackage{placeins}
\usepackage{float}
\usepackage{url}

\def\tsc#1{\csdef{#1}{\textsc{\lowercase{#1}}\xspace}}
\tsc{WGM}
\tsc{QE}

\begin{document}
\let\WriteBookmarks\relax

\shorttitle{CASA-SDF: Curriculum-Aware Spatial Adaptation}    

\shortauthors{L. Yang et al.}  

\title [mode = title]{CASA-SDF: Curriculum-Aware Spatial Adaptation with Curvature-Guided Density for Neural Implicit Surface Reconstruction}

\author[1]{Lei Yang}
\ead{jubary@njust.edu.cn}
\credit{Conceptualization, Methodology, Software, Validation, Investigation, Visualization, Writing - original draft}

\author[1]{Weiqing Li}[orcid=0000-0002-1929-3654]
\ead{li_weiqing@njust.edu.cn}
\credit{Conceptualization, Formal analysis, Supervision, Writing - review \& editing}

\author[2]{Zhiyong Su}[orcid=0000-0003-1361-8451]
\ead{suzhiyong@njust.edu.cn}
\credit{Conceptualization, Supervision, Writing - review \& editing}

\author[1]{Liang Xiao}[orcid=0000-0003-0178-9384]
\cormark[1]
\ead{xiaoliang@njust.edu.cn}
\credit{Conceptualization, Resources, Funding acquisition, Supervision, Project administration, Writing - review \& editing}
\cortext[cor1]{Corresponding author}

\affiliation[1]{organization={School of Computer Science and Engineering, Nanjing University of Science and Technology},
  addressline={200 Xiaolingwei},
  city={Nanjing},
  state={Jiangsu},
  country={China},
  postcode={210094}
}
\affiliation[2]{organization={School of Automation, Nanjing University of Science and Technology},
  addressline={200 Xiaolingwei},
  city={Nanjing},
  state={Jiangsu},
  country={China},
  postcode={210094}
}

\begin{abstract}
  Neural implicit representations have emerged as a powerful paradigm for 3D reconstruction. However, high-fidelity indoor surface reconstruction remains a significant challenge, primarily due to the pronounced \emph{geometric heterogeneity} of indoor scenes. Large texture-less planar regions typically require stronger regularization to suppress high-frequency artifacts, while thin structures demand sharper, more adaptive representations to mitigate the spectral bias of multi-layer perceptrons (MLPs) and prevent over-smoothing. Existing approaches often rely on spatially indiscriminate prior supervision and a scene-global SDF-to-density transformation, which constrains their ability to balance planar smoothness and detail preservation. In this paper, we propose CASA-SDF (Curriculum-Aware Spatial Adaptation for SDF), a unified framework that addresses this challenge via complementary adaptations of supervision and representation capacity. Specifically, Hybrid Spatially-Adaptive Uncertainty Annealing (SAUA) fuses semantic and photometric uncertainties to construct a pixel-wise curriculum for monocular prior supervision. This strategy maintains regularization in reliable regions while attenuating unreliable supervision early in training to enable data-driven photometric refinement. Meanwhile, Curvature-Aware Locally Adaptive Density Transformation (CALADT) progressively modulates the sharpness of the SDF-to-density mapping via a curvature proxy to enhance the representation of thin structures. Extensive experiments on benchmark indoor datasets demonstrate that CASA-SDF improves surface completeness and detail recovery on high-frequency structures, without compromising the stability of planar surfaces.
\end{abstract}


\maketitle

\vspace{0.6em}
\begin{center}
\footnotesize
\textcopyright{} 2026. This manuscript version is made available under the CC-BY-NC-ND 4.0 license. This is the accepted version of the article published in Neurocomputing, available online at: \url{https://doi.org/10.1016/j.neucom.2026.134488}
\end{center}
\vspace{1.2em}

\section{Introduction}
\label{sec:intro}

\begin{figure*}[t]
  \centering
  \includegraphics[width=\textwidth]{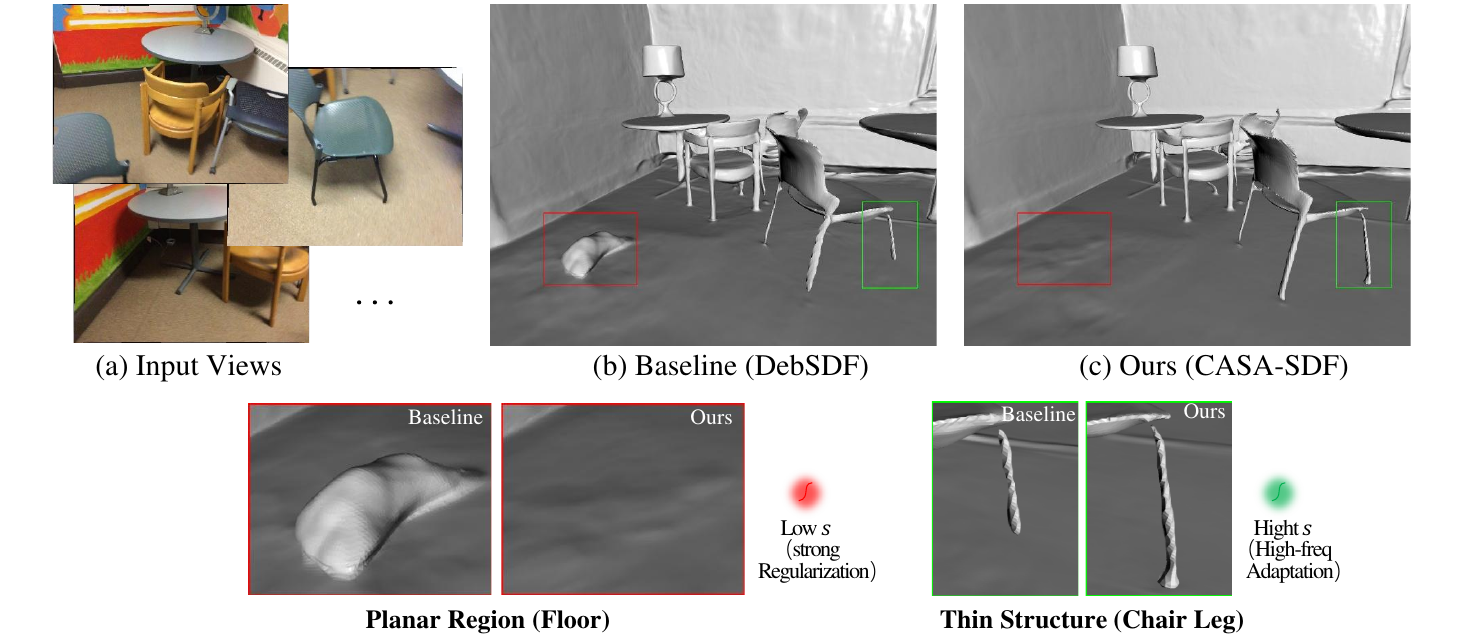}
  \caption{Breaking the trade-off in indoor reconstruction via curriculum-aware spatial adaptation.
  Indoor scenes exhibit \emph{geometric heterogeneity}, which creates a fundamental conflict for standard reconstruction methods.
  \textbf{(a)} Input Views.
  \textbf{(b)} A strong baseline (e.g., DebSDF~\cite{Debsdf}) suffers from rigid global regularization scheme, resulting in noisy artifacts on texture-less planar regions (red box) and over-smoothed, broken geometry on thin structures (green box).
  \textbf{(c)} CASA-SDF alleviates this conflict. By adapting both supervision and density representation, it produces smooth wall reconstructions (bottom-left zoom-in, corresponding to low $s$) while preserving sharp high-frequency details on chair legs (bottom-right zoom-in, corresponding to high $s$).}
  \label{fig:teaser}
\end{figure*}

Learning continuous neural implicit representations from calibrated multi-view images has established itself as a dominant paradigm for 3D scene reconstruction, effectively bridging the gap between deep representation learning and differentiable rendering~\cite{Nerf,NeuS,volsdf}. This line of research has further expanded to generalizable and sparse-view scenarios~\cite{PixelNeRF,MVSNeRF,SparseNeuS,Sparis}, highlighting the versatility of implicit neural fields. However, high-fidelity reconstruction of complex indoor scenes remains a particular challenge. Indoor environments exhibit pronounced \emph{geometric heterogeneity}, characterized by the coexistence of low-frequency planar regions and high-frequency thin structures. In practice, expansive texture-less surfaces (e.g., walls and floors) coexist with slender objects and sharp boundaries (e.g., chair legs and furniture edges). This geometric diversity gives rise to a fundamental optimization dilemma: planar regions benefit from strong regularization to suppress neural noise, whereas thin structures require higher-frequency adaptivity to avoid over-smoothing (Fig.~\ref{fig:teaser}).

Recent indoor reconstruction methods have markedly improved geometry quality by incorporating monocular priors such as depth and normals~\cite{MonoSDF,NeuRIS,PSDF,Sun_2024,PMVC}. MonoSDF~\cite{MonoSDF}, in particular, demonstrates the practical value of such priors for stabilizing neural implicit reconstruction in texture-less scenes. More recent prior-aware and bias-aware methods further recognize that these priors are not uniformly reliable across space. DebSDF~\cite{Debsdf} shows that uncertainty and rendering bias should be considered during optimization, while ND-SDF~\cite{ndsdf} demonstrates that correcting prior--scene normal mismatch can further improve high-fidelity indoor reconstruction. However, despite these advances, existing indoor SDF methods still mainly improve prior reliability, prior correction, or bias reduction in isolation, and therefore do not yet fully address the coupled effects of geometric heterogeneity within a single scene.

A first unresolved challenge concerns \emph{adaptation of the supervision lifecycle}. While monocular priors provide reliable guidance on dominant planar regions, they degrade near boundaries, thin structures, and appearance-ambiguous areas. Crucially, their supervisory value evolves over training: a prior instrumental for coarse stabilization early on may become restrictive once the neural field accumulates sufficient multi-view evidence for finer geometric refinement. Existing strategies---relying on static weighting, hard masking, or scene-global prior handling---overlook this temporal evolution in supervisory value.

A second, complementary challenge concerns \emph{representation bandwidth adaptation}. NeuS~\cite{NeuS} and VolSDF~\cite{volsdf} employ a globally shared sharpness parameter $s$---equivalently bandwidth $\beta = 1/s$---within the SDF-to-density transformation. This global parameterization imposes an inherent trade-off: small $s$ suppresses noise on planar surfaces yet blurs thin structures, whereas large $s$ preserves high-frequency detail at the cost of amplified noise and instability in smooth regions. Existing indoor methods predominantly improve supervision quality or correct prior bias, yet neglect the spatial heterogeneity of bandwidth demand within the SDF-to-density mapping.

In this paper, we propose CASA-SDF (Curriculum-Aware Spatial Adaptation for SDF), a unified framework that treats indoor geometric heterogeneity as a coupled problem of \emph{supervision lifecycle adaptation} and \emph{representation bandwidth adaptation}. The underlying observation is that unreliable priors and insufficient local bandwidth arise from the same scene heterogeneity, but affect optimization through different mechanisms: large planar regions benefit from sustained regularization, whereas thin structures and sharp boundaries require both earlier relaxation of unreliable priors and sharper local surface localization. 

To address the first aspect, we introduce Hybrid Spatially-Adaptive Uncertainty Annealing (SAUA), a pixel-wise curriculum that fuses semantic and photometric uncertainties to schedule monocular prior supervision throughout training. SAUA preserves regularization in confident planar regions while rapidly annealing prior influence in uncertain boundary and thin-structure regions, thereby allowing photometric evidence to play a stronger role as optimization proceeds. To address the second aspect, we introduce Curvature-Aware Locally Adaptive Density Transformation (CALADT), which modulates the sharpness parameter $s(\mathbf{x}, \tau)$ using a curvature-magnitude proxy and training progress. CALADT allocates higher bandwidth to geometrically complex regions while maintaining smoother behavior on planar regions, improving thin-structure recovery without compromising the overall stability of the neural field. Together, these two mechanisms provide a unified spatial-temporal adaptation of supervision and representation bandwidth for indoor neural surface reconstruction.

Our contributions are summarized as follows:
\begin{itemize}
    \item We formulate indoor geometric heterogeneity as a coupled problem of supervision-lifecycle adaptation and representation-bandwidth adaptation, providing a unified perspective on reconciling planar smoothness and thin-structure preservation.
    \item We propose Hybrid-SAUA, an uncertainty-driven mechanism that conservatively fuses semantic and photometric uncertainties to prevent erroneous monocular constraints from degrading delicate geometry.
    \item We propose CALADT, a curvature-aware locally adaptive sharpness modulation with progressive activation, enabling spatially varying representation bandwidth for precise thin-structure localization.
    \item Extensive experiments on ScanNet and Replica demonstrate that CASA-SDF improves surface completeness and recall on high-frequency structures and transitional regions, with competitive quality maintained on planar surfaces and in global metrics.
\end{itemize}

\section{Related Work}
\label{sec:related_work}
\subsection{Neural Implicit Surface Reconstruction}
Neural implicit representations have reshaped multi-view reconstruction, owing to their continuous formulation and differentiable rendering capabilities. NeRF~\cite{Nerf} learns a volumetric radiance field via volume rendering, but extracting high-quality surfaces from density fields remains non-trivial. To explicitly model geometry, SDF/occupancy-based formulations have been introduced for surface reconstruction~\cite{IDR_2020,Differentiable}. Among them, NeuS~\cite{NeuS} and VolSDF~\cite{volsdf} establish SDF-induced density transformations that enable surface extraction from the zero-level set while retaining the robustness of volume rendering.
Recent works have focused on enhancing fidelity, scalability, and robustness. High-fidelity surface reconstruction has been advanced through more robust neural surface pipelines~\cite{neuralangelo} and geometry-oriented improvements~\cite{HF-neus}. Efficiency and scalability have been improved via structured or hybrid representations, such as voxel-based acceleration and lattice/hash-style parameterizations~\cite{Voxurf,PermutoSDF,BakedSDF}. Extensions to broader scene scenarios have also been explored, e.g., street-scale implicit surfaces~\cite{StreetSurf}. For texture-less regions and weak photometric cues, structural similarity objectives have proven effective for neural fields~\cite{S3IM}.
Recent indoor-focused methods further emphasize recovering fine details through hybrid representations, enhanced priors~\cite{Fine-Recon}, or auxiliary-field augmentation of the SDF formulation for higher-fidelity indoor geometry~\cite{ndsdf}. A persistent challenge in indoor scenes is that global rendering and parameterization choices tend to be suboptimal under geometric heterogeneity. DebSDF~\cite{Debsdf} analyzes bias and failure modes in NeuS-style indoor reconstruction and enhances robustness via objective-level correction and bias-aware SDF-to-density design. Other works address indoor-specific appearance issues (e.g., reflections) through ambiguity reduction and view-dependent compensation~\cite{Ref-NeuS,NC-SDF}. Complementary efforts improve optimization and sampling for neural implicit surface rendering, including neural importance sampling and probability-guided sampling~\cite{NeuSample,Pais_2024}, as well as rendering variants that stabilize supervision~\cite{RayDistance}.
\subsection{Indoor Scene Reconstruction with Geometric Priors}
Indoor reconstruction remains particularly challenging due to extensive texture-less planar regions and repeated patterns, where photometric constraints alone are insufficiently informative. A long-standing research direction has been to constrain reconstruction using structural assumptions, such as Manhattan-world priors~\cite{Manhattansdf}, and modern variants further integrate indoor planar priors explicitly~\cite{IPP}.
Recent SDF-based indoor reconstruction methods commonly incorporate dense monocular priors (depth/normal) predicted by pre-trained networks on multi-task datasets~\cite{Omnidata}, which effectively improves reconstruction stability and quality~\cite{MonoSDF,NeuRIS}. Prior-driven implicit surface learning has also been explored for multi-view reconstruction under limited observation conditions~\cite{PSDF}, and more recent work introduces stronger indoor consistency priors for few-shot settings~\cite{Sun_2024} or promotes multi-view consistency explicitly~\cite{PMVC}.
The field is also shifting toward stronger priors derived from large-scale models. Diffusion-based and foundation-style monocular depth priors enhance zero-shot generalization and provide more reliable supervision in diverse indoor scenes~\cite{marigold,depthanythingv2}. Generative priors are further explored for image-to-3D and view synthesis pipelines~\cite{Wonder3D,WildFusion}, while radiance-field priors can guide indoor reconstruction~\cite{nerfprior}. Related prior-guided reconstruction has also been explored beyond indoor scenes, e.g., satellite imagery and in-the-wild settings~\cite{Sat-DN,GeoPriorWild}. Sparse-view indoor implicit surface reconstruction remains an active topic~\cite{SparseNeuS,Sparis}, highlighting the ongoing need for robust prior integration and adaptive optimization.
\subsection{Uncertainty and Reliability in Neural Fields}
Quantifying uncertainty is critical for robust reconstruction, particularly when supervision sources exhibit heterogeneous reliability across space. Probabilistic formulations have been proposed to capture predictive uncertainty within neural fields~\cite{StochasticNeRF,CF-NeRF}, and uncertainty has also been used for active view selection and planning~\cite{ActiveNeRF,ActiveRMAP}. More recent approaches estimate reliability using information-theoretic signals such as Fisher information~\cite{FisherRF}, or through SDF-specific approximations for uncertainty estimation~\cite{BayesSDF}.
In indoor reconstruction, uncertainty is frequently leveraged to mitigate the effects of unreliable monocular priors and stabilize optimization~\cite{Debsdf,RegSDF}. Our SAUA is related to this line of research but differs in two key aspects. First, it efficiently derives semantic uncertainty from an Angular von Mises–Fisher estimator~\cite{EEAU} without computationally expensive approximations. Second, it fuses semantic uncertainty with photometric uncertainty from multi-view consistency (via patch warping)~\cite{NeuralWarp}, and uses the fused reliability to drive a pixel-wise curriculum schedule, rather than relying on static weighting or hard filtering.
\paragraph{Relation to Prior-Aware Indoor SDF Methods.}
Recent indoor neural reconstruction has converged on three practical findings: monocular priors are useful for indoor geometry recovery~\cite{MonoSDF}, uncertainty and bias awareness help when those priors become unreliable~\cite{Debsdf}, and explicit prior correction can improve reconstruction fidelity~\cite{ndsdf}. CASA-SDF builds on these advances while addressing a complementary dimension. Rather than introducing an additional prior source or another direct prior-correction field, it addresses the coupled effects of geometric heterogeneity on two aspects of the reconstruction process: \emph{how long} priors should guide optimization and \emph{where} the implicit representation should allocate higher bandwidth. While prior-aware and bias-aware methods focus on improving supervision reliability, debiasing, or prior--scene consistency, CASA-SDF complements these by modeling the spatially varying bandwidth demand of the SDF representation itself. SAUA and CALADT therefore operate on complementary aspects of the same problem, jointly addressing indoor geometric heterogeneity through unified supervision and representation adaptation.

\section{Our Method}
\label{sec:method}
\begin{figure*}[t]
\centering
\IfFileExists{figures/pipeline.pdf}{\includegraphics[width=\textwidth,height=0.8\textwidth]{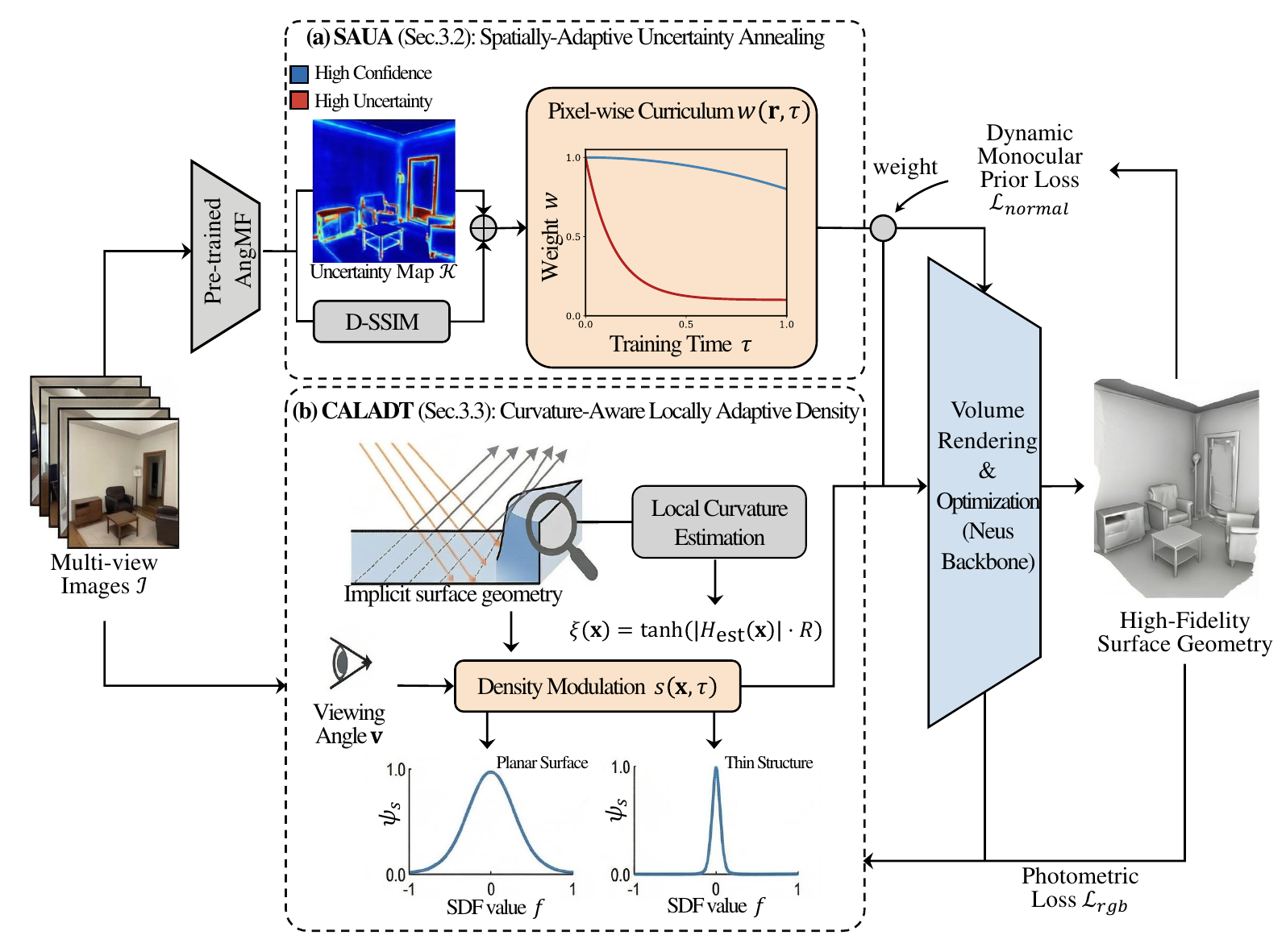}}{}
\caption{Overview of CASA-SDF. The pipeline consists of two complementary mechanisms. \textbf{(a)} SAUA (Sec.~\ref{sec:saua}): a pre-trained Angular von Mises--Fisher (AngMF) estimator estimates aleatoric uncertainty $\kappa$, which drives a pixel-wise curriculum for annealing monocular prior supervision. Reliable regions (e.g., walls) retain regularization, whereas unreliable regions (e.g., edges) transition earlier to photometric refinement. \textbf{(b)} CALADT (Sec.~\ref{sec:caladt}): the sharpness parameter $s$ is modulated by a curvature-magnitude proxy with a progressive schedule, yielding sharper density peaks on thin structures and smoother distributions on planar surfaces while maintaining stable surface localization in practice.}
\label{fig:pipeline}
\end{figure*}

Given a set of calibrated multi-view images $\mathcal{I} = \{I_i\}_{i=1}^M$, our goal is to reconstruct high-fidelity indoor surface geometry. As discussed in Sec.~\ref{sec:intro}, implicit indoor scene reconstruction faces a fundamental conflict under \textit{geometric heterogeneity}: planar regions require strong regularization to suppress neural noise, whereas thin structures require higher-frequency adaptivity to overcome the spectral bias of MLPs. To resolve this dilemma, we propose CASA-SDF (Curriculum-Aware Spatial Adaptation for SDF), a curriculum-guided representation learning framework. Our method comprises two synergistic components (Fig.~\ref{fig:pipeline}): 
    (1) Hybrid Spatially-Adaptive Uncertainty Annealing (SAUA) (Sec.~\ref{sec:saua}), which formulates an uncertainty-driven curriculum learning strategy to dynamically anneal prior supervision by fusing semantic and photometric uncertainties; 
    (2) Curvature-Aware Locally Adaptive Density Transformation (CALADT) (Sec.~\ref{sec:caladt}), which adaptively modulates the neural activation mapping (sharpness parameter) based on local curvature and training progress to resolve the bias-variance trade-off in the neural implicit representation.
\subsection{Preliminaries and Problem Formulation}
\label{sec:preliminaries}
We adopt the implicit surface representation from NeuS~\cite{NeuS}. The scene geometry is represented as a Signed Distance Field (SDF) $f_\theta: \mathbb{R}^3 \to \mathbb{R}$, and appearance is modeled by a radiance field $\mathbf{c}_\phi: \mathbb{R}^3 \times \mathbb{S}^2 \to \mathbb{R}^3$, both parameterized by multi-layer perceptrons (MLPs).
For a camera ray $\mathbf{r}(t) = \mathbf{o} + t\mathbf{d}$, we denote the SDF value along the ray as $f(t)=f_\theta(\mathbf{r}(t))$. Following NeuS~\cite{NeuS} and HF-NeuS~\cite{HF-neus}, we model a CDF-based transparency/occupancy surrogate as a transformed SDF, defined as:
\begin{equation}
T(t)=\Psi_{s}(f(t))=\frac{1}{1+\exp(-s\,f(t))},
\end{equation}
where we follow the NeuS parameterization in which $s>0$ is an inverse scale (sharpness) controlling the transition near the surface. Note that $T(t)$ here denotes the CDF-based surrogate used in NeuS, rather than the standard accumulated transmittance in NeRF. NeuS uses this signed-distance CDF to derive differentiable, non-negative rendering weights. The corresponding induced weighting density is derived from this surrogate:
\begin{equation}
\rho(t) = -\frac{d}{dt}\log T(t),
\end{equation}
and the volume rendering weight is $w(t)=T(t)\rho(t)$. The expected color is then computed by
\begin{equation}
C(\mathbf{r}) = \int_{0}^{\infty} w(t)\,\mathbf{c}(\mathbf{r}(t), \mathbf{d}) \, dt.
\end{equation}
However, standard methods typically employ a globally shared sharpness parameter $s$ for the entire scene. A constant sharpness forces a trade-off: small $s$ (smooth) is preferred for planar regions to suppress noise, while large $s$ (sharp) is required for thin structures to ensure precise high-frequency localization. While DebSDF~\cite{Debsdf} mitigates bias in indoor SDF reconstruction via bias-aware objective design, it still largely assumes a scene-global sharpness.
\subsection{Spatially-Adaptive Uncertainty Annealing}
\label{sec:saua}
Existing methods~\cite{MonoSDF,NeuRIS} typically incorporate monocular priors (normal/depth) with static weights or simple thresholding. However, priors exhibit distinct failure modes: CNN-based priors struggle with high-frequency edges (semantic uncertainty), while multi-view stereo (MVS) cues fail in texture-less regions (photometric uncertainty)---consistent with the challenges of indoor geometric heterogeneity discussed earlier.
We propose Spatially-Adaptive Uncertainty Annealing (SAUA) to fuse these complementary uncertainties into a unified curriculum for uncertainty-aware prior scheduling.

\begin{figure}
\centering
\includegraphics[width=\columnwidth]{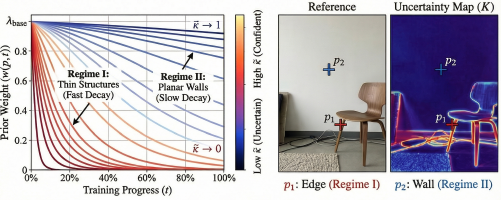}
  \caption{\textbf{Visualization of the SAUA curriculum.} Unlike static weighting, the proposed pixel-wise weights follow different trajectories over training. High-uncertainty regions (red) are annealed rapidly to release unreliable constraints, whereas confident regions (blue) retain regularization for a longer period.}
  \label{fig:annealing}
\end{figure}

We explicitly model two types of uncertainty:
\begin{itemize}
\item \textbf{Semantic uncertainty ($\mathcal{U}_{\text{sem}}$):}
We employ an Angular von Mises--Fisher (AngMF) estimator~\cite{EEAU} to predict pixel-wise normal confidence. Let $\kappa \in \mathbb{R}^+$ be the concentration parameter predicted by the estimator. We define the normalized semantic uncertainty as:
\begin{equation}
\mathcal{U}_{\text{sem}}(\mathbf{r}) = \exp\left(-\frac{\kappa(\mathbf{r})}{\kappa_0}\right),
\end{equation}
where $\kappa_0$ is a normalization constant introduced to mitigate scale-induced distribution shifts of $\mathcal{U}_{\text{sem}}$ across different scenes. It is computed once before optimization using the predictions from the frozen AngMF estimator. Specifically, we set $\kappa_0$ as the median of $\kappa$ over all training pixels to ensure robustness against extreme outliers, and it is kept fixed during training. This formulation maps $\kappa\in[0,\infty)$ to $\mathcal{U}_{\text{sem}}\in(0,1]$, where high-confidence regions (e.g., planar walls) yield $\mathcal{U}_{\text{sem}} \to 0$, while ambiguous structures (e.g., edges) result in high uncertainty, eliminating the need for manual thresholding.
\item \textbf{Photometric uncertainty ($\mathcal{U}_{\text{photo}}$):}
To identify regions where multi-view consistency is unreliable (e.g., occlusion and view-dependent appearance), we measure patch-level structural dissimilarity across views via $\text{DSIM}=1-\text{SSIM}$ on warped patches~\cite{NeuralWarp} and aggregate it with a best-match strategy:
\begin{equation}
\mathcal{U}_{\text{photo}}(\mathbf{r}) = 1 - \max_{v \in \mathcal{V}} \text{SSIM}\left(P_{\text{ref}}, P_{v \to \text{ref}}(\hat{D}(\mathbf{r}))\right),
\end{equation}
where $\mathcal{V}$ denotes the set of neighboring source views selected based on co-visibility, $P_{\text{ref}}$ is the patch centered at the ray's intersection in the reference view, and $P_{v \to \text{ref}}$ is the corresponding patch from view $v$ warped to the reference perspective using the monocular prior depth $\hat{D}(\mathbf{r})$ and known camera geometry. The max operator achieves a robust best-match aggregation to reduce the impact of occlusions. If no valid source view is available, we set 
$\mathcal{U}_{\text{photo}}(\mathbf{r})=1$ to indicate maximum uncertainty.
\end{itemize}
\paragraph{Uncertainty-Aware Curriculum.}
We integrate these metrics via a conservative max-based fusion strategy:
\begin{equation}
    \mathcal{U}_{\text{total}}(\mathbf{r}) = \max \big( \mathcal{U}_{\text{sem}}(\mathbf{r}), \mathcal{U}_{\text{photo}}(\mathbf{r}) \big),
\end{equation}
which ensures that prior supervision is reduced if \textit{either} source indicates unreliability.
To implement this uncertainty-driven learning strategy, we define the time-dependent prior weight $w_{\text{prior}}(\mathbf{r}, \tau)$ for a ray $\mathbf{r}$ at training progress $\tau \in [0,1]$ as:
\begin{equation}
w_{\text{prior}}(\mathbf{r}, \tau) = \lambda_{\text{base}} \cdot \exp\left( - \frac{\eta \cdot \tau}{1 - \mathcal{U}_{\text{total}}(\mathbf{r}) + \epsilon} \right),
\label{eq:saua}
\end{equation}
where $\lambda_{\text{base}}$ represents the initial regularization strength, $\eta$ is a hyperparameter controlling the global annealing speed, and $\epsilon$ is a small constant for numerical stability.
This formulation induces distinct optimization dynamics that differentiate the role of supervision over time:
\begin{itemize}
    \item Regime I (High Confidence, $\mathcal{U}_{\text{total}} \to 0$): For reliable regions like planar walls, the denominator $(1 - \mathcal{U}_{\text{total}} + \epsilon)$ remains close to $1$. The prior weight undergoes a relatively gentle and delayed decay. It retains supervisory influence significantly longer during the optimization process, effectively suppressing the high-frequency noise on texture-less surfaces in neural implicit reconstruction.
    \item Regime II (High Uncertainty, $\mathcal{U}_{\text{total}} \to 1$): For ambiguous regions such as thin chair legs and other high-frequency structures, the denominator approaches $\epsilon$, causing a rapid exponential decay in $w_{\text{prior}}(\mathbf{r}, \tau)$. The constraint vanishes early in training, preventing the neural field from being dominated by biased priors and allowing the multi-view photometric loss ($\mathcal{L}_{\text{rgb}}$) to drive data-driven fine-grained refinement.
\end{itemize}
This uncertainty-aware formulation therefore yields region-dependent decay trajectories: reliable regions remain regularized for longer, whereas uncertain regions are released earlier for photometric refinement. Fig.~\ref{fig:annealing} visualizes these distinct annealing behaviors.
\subsection{Curvature-Aware Locally Adaptive Density Transformation}
\label{sec:caladt}
Standard implicit surface models employ a globally shared sharpness parameter $s$ as the neural activation mapping that transforms the SDF into a density field. 
This rigid mapping creates a representation trade-off that closely mirrors the optimization dilemma induced by indoor geometric heterogeneity: a small $s$ (smooth) is desirable on planar regions to regularize the neural field and suppress noise, whereas a large $s$ (sharp) is needed to provide sufficient representation capacity for thin structures and precise high-frequency localization. Unlike prior works that modify the SDF field directly, our design adapts the SDF-to-density transformation itself. This progressive schedule mitigates potential peak shift in the induced weighting kernel by delaying spatial modulation until curvature estimates stabilize.
\paragraph{Local Curvature Estimation.}
We characterize geometric complexity via a normal-variation based curvature-magnitude proxy. Intuitively, regions with rapid normal variation require sharper surface localization (thin structures, edges), while planar regions naturally yield a near-zero response. For a smooth surface, the normal variation along a tangent direction is related to the shape operator, so $\|\mathbf{n}(\mathbf{x}+\delta\mathbf{t})-\mathbf{n}(\mathbf{x})\|/\delta$ provides an efficient proxy of local curvature magnitude. To avoid the computational cost of second-order derivatives, we compute this proxy via finite differences of normals on the tangent plane. For a query point $\mathbf{x}$, we sample orthogonal tangent vectors $\{\mathbf{t}_1, \mathbf{t}_2\}$:
\begin{equation}
|H_{\text{est}}(\mathbf{x})| \approx \frac{1}{2\delta} \sum_{i=1}^{2} \| \mathbf{n}(\mathbf{x}) - \mathbf{n}(\mathbf{x} + \delta \mathbf{t}_i) \|_2,
\end{equation}
where $\delta = 0.01 \cdot R$ is the step size relative to the scene bounding sphere radius $R$. 
This normalization ensures scale-invariant curvature estimation, thereby 
preserving hyperparameter consistency across datasets.
Since this is not the exact mean curvature $|H(\mathbf{x})|$ but a finite-difference approximation, we treat $|H_{\text{est}}(\mathbf{x})|$ as a curvature complexity proxy and map it to a normalized modulation factor $\xi(\mathbf{x}) \in [0, 1]$ using a hyperbolic tangent compression:
\begin{equation}
\xi(\mathbf{x}) = \tanh(|H_{\text{est}}(\mathbf{x})| \cdot R).
\end{equation}
\paragraph{Adaptive Sharpness Modulation Definition.}
We define the spatially-varying sharpness $s(\mathbf{x}, \tau)$ as an increasing function of the geometric complexity:
\begin{equation}
s(\mathbf{x}, \tau) = s_{\text{base}}\left(1 + g(\tau) \cdot \lambda_H \cdot \xi(\mathbf{x})^{\alpha_H}\right)
\label{eq:adaptive_s},
\end{equation}
where $s_{\text{base}}$ denotes the global base sharpness. High-curvature regions (large $\xi$) are assigned larger $s$, producing sharper surface localization, while planar regions retain smaller $s$ to suppress noise. We denote the curvature-aware modulation term as $\Psi_H(\mathbf{x},\tau)=1 + g(\tau) \cdot \lambda_H \cdot \xi(\mathbf{x})^{\alpha_H}$.
\paragraph{Progressive Activation.}
To prevent premature modulation when curvature estimates are unreliable, we introduce a progress gate $g(\tau)$ implemented as a piecewise-linear ramp: $g(\tau)=0$ for $\tau\le\tau_0$, increases linearly to $1$ for $\tau\in[\tau_0,\tau_1]$, and remains $1$ for $\tau\ge\tau_1$. This gate delays spatial modulation until the geometry and curvature estimates become sufficiently reliable.
\paragraph{Properties.}
Notably, CALADT degenerates to a standard global-sharpness NeuS/HF-NeuS formulation when $\lambda_H=0$ (or $g(\tau)=0$), ensuring interpretability of the added hyperparameters and backward compatibility with existing frameworks.
\paragraph{Implementation Details.}
With CALADT, we substitute the spatially varying sharpness into the CDF-based surrogate along each ray:
\begin{equation}
T(t)=\Psi_{s(\mathbf{r}(t),\tau)}(f(t)).
\end{equation}
We follow NeuS to compute differentiable, non-negative rendering weights, where $s$ is evaluated at each sampled point $\mathbf{r}(t)$ and substituted into the weight computation without explicitly expanding the chain rule in $\rho(t)$ under spatially varying $s$. Following the NeuS/HF-NeuS implementation, we evaluate $s(\mathbf{r}(t), \tau)$ at each sampled point but treat it as locally constant when computing the density derivative $\rho(t) = -\frac{d}{dt}\log T(t)$. Under this approximation, the weight computation retains the standard NeuS form at each sample, and the gradient of $s$ with respect to position does not propagate through the density derivative.
\paragraph{Remarks on Localization Fidelity.}
Spatially varying sharpness introduces a position-dependent scaling of the SDF that can, in principle, shift the peak of the induced weighting kernel $w(t)$ if the spatial gradient $\nabla s(\mathbf{x}, \tau)$ is fully back-propagated through the density derivative. To maintain stable surface localization, we employ a practical approximation: the sharpness modulation term $\Psi_H(\mathbf{x},\tau)$ is treated as locally constant during the computation of $\rho(t)$, effectively applying a stop-gradient to the curvature-dependent component. Under this approximation, the weight computation retains the symmetric kernel structure of NeuS, and no noticeable peak drift is observed in practice (see the Supplementary Material for a 1D illustration).

\subsection{Optimization Objective}
We train the network end-to-end using a composite objective combining photometric and geometric terms:
\begin{align}
\mathcal{L} &= \mathcal{L}_{\text{rgb}} + \lambda_{\text{eik}} \mathcal{L}_{\text{eik}} \nonumber\\
    &\quad + \sum_{\mathbf{r} \in \mathcal{R}} w_{\text{prior}}(\mathbf{r}, \tau) \left( \lambda_{n} \mathcal{L}_{\text{normal}}(\mathbf{r}) + \lambda_{d} \mathcal{L}_{\text{depth}}(\mathbf{r}) \right).
\end{align}
The photometric loss $\mathcal{L}_{\text{rgb}}$ is the L1 error between rendered and ground-truth colors.
A key feature of this objective is that geometric priors are dynamically gated by our SAUA weights $w_{\text{prior}}(\mathbf{r}, \tau)$ derived in Sec.~\ref{sec:saua}:
\begin{align}
\mathcal{L}_{\text{normal}}(\mathbf{r}) &= \| 1 - \hat{\mathbf{n}}(\mathbf{r})^\top \mathbf{N}(\mathbf{r}) \|_1, \\
    \mathcal{L}_{\text{depth}}(\mathbf{r}) &= \| \hat{D}(\mathbf{r}) - D(\mathbf{r}) \|_1.
\end{align}
Here, $\hat{\mathbf{n}}$ and $\hat{D}$ are monocular priors, while $\mathbf{N}$ and $D$ are rendered quantities. Note that monocular depth $\hat{D}$ is aligned to the metric scale via median scaling. This unified objective ensures that reliable monocular priors regularize the geometry, while unreliable ones (attenuated by $w_{\text{prior}}(\mathbf{r}, \tau)$) are progressively discarded to allow data-driven refinement.
\FloatBarrier

\section{Experiments}
\label{sec:experiments}

We evaluate CASA-SDF on challenging indoor scenes and analyze the contributions of the proposed external adaptation (SAUA) and internal adaptation (CALADT) to improved reconstruction quality. Given our focus on \textit{geometric heterogeneity}, we emphasize two consistently challenging regimes for indoor reconstruction: (i) thin structures and boundaries, where monocular priors are often unreliable; and (ii) large texture-less planar regions, where photometric constraints are weakly informative.

\subsection{Experimental Setup}

\noindent\textbf{Datasets.}
We evaluate CASA-SDF on two standard indoor datasets: \textbf{ScanNet}~\cite{scanNet} and \textbf{Replica}~\cite{Replica}. ScanNet comprises large-scale real-world RGB-D indoor scenes with complex geometry and extensive texture-less regions; following MonoSDF~\cite{MonoSDF}, we conduct evaluations on 8 representative scenes. Replica is a high-fidelity synthetic dataset with clean ground-truth geometry, where we use 8 scenes to assess reconstruction accuracy under controlled conditions.

\noindent\textbf{Baselines.}
We compare CASA-SDF with representative methods categorized into three groups: (i) \textit{Traditional MVS}: COLMAP~\cite{colmap}; (ii) \textit{Neural surface reconstruction without explicit monocular priors}: 
NeuS~\cite{NeuS} and VolSDF~\cite{volsdf} (included for reference, though 
their standard configurations are not optimized for indoor scenes);
(iii) \textit{Neural reconstruction with priors / indoor-specific designs}: 
Manhattan-SDF~\cite{Manhattansdf}, NeuRIS~\cite{NeuRIS}, MonoSDF~\cite{MonoSDF}, 
HelixSurf~\cite{HelixSurf}, Fine-Recon~\cite{Fine-Recon}, DebSDF~\cite{Debsdf}, 
ND-SDF~\cite{ndsdf}, and occSDF~\cite{occsdf}.

For a comprehensive evaluation, we adopt a two-level comparison protocol. The broad benchmark tables aggregate results from prior publications or official implementations evaluated under their established protocols. Additionally, we reproduce controlled results for the strongest prior-aware and bias-aware baselines most relevant to our claims (MonoSDF, DebSDF, and ND-SDF) using identical scene splits, mesh extraction resolutions, culling strategies, and evaluation scripts. This controlled reproduction minimizes cross-implementation ambiguity and provides a standardized setting for isolating algorithmic improvements.

\noindent\textbf{Metrics.}
Following established indoor reconstruction protocols~\cite{MonoSDF,Debsdf}, we report Accuracy (Acc$\downarrow$), Completeness (Comp$\downarrow$), Precision (Prec$\uparrow$), Recall (Rec$\uparrow$), and F-score ($\uparrow$) with a 5\,cm evaluation threshold. For ablation studies, we additionally report the compact Acc/Comp/F-score triplet for improved readability. For region-aware analysis, we use the same 5\,cm evaluation threshold and compute region-specific metrics under fixed geometric partitions defined on the ground-truth mesh.

\subsection{Implementation Details}

\noindent\textbf{Network Architecture.}
Our geometry network is an 8-layer MLP with 256 hidden units and Softplus activation, which predicts the SDF value and a 256-dimensional feature vector. The color network is a 4-layer MLP that takes the feature vector, 3D position, and viewing direction as input. We apply positional encoding with 6 frequencies for spatial coordinates and 4 frequencies for viewing directions, following the setup in NeuS~\cite{NeuS}.

\noindent\textbf{Curriculum-Aware Configuration (SAUA).}
For SAUA (Eq.~\ref{eq:saua}), we use a pre-trained AngMF estimator~\cite{EEAU} to obtain pixel-wise uncertainty (via $\kappa$), which is kept frozen during optimization. We set $\lambda_{\text{base}} = 0.1$, the global annealing rate $\eta = 4.0$, and $\epsilon = 10^{-3}$.

\noindent\textbf{Density Modulation Settings (CALADT).}
For CALADT, we use the spatially-varying sharpness parameter $s(\mathbf{x},\tau)$ (with bandwidth $\beta(\mathbf{x},\tau)=1/s(\mathbf{x},\tau)$ in some formulations). We set the base sharpness $s_{\text{base}}=1/\beta_0$ with $\beta_0=0.3$. The finite-difference step size for curvature estimation follows Sec.~\ref{sec:caladt}, i.e., $\delta = 0.01 \cdot R$. For the curvature modulation term $\Psi_H$, we set $\lambda_H = 1.0$ and $\alpha_H = 2.0$. We use $\tau_0=0.3$ and $\tau_1=0.6$ for the progressive gate $g(\tau)$. To control computational cost, curvature estimation is activated only for samples near the surface (e.g., $|f(\mathbf{x})| < 5\,\mathrm{cm}$).

\noindent\textbf{Training Protocol.}
We implement CASA-SDF in PyTorch and train each scene on a single NVIDIA RTX 4090 GPU. The learning rate is warmed up to $5 \times 10^{-4}$ during the first 5{,}000 iterations and subsequently decayed via a cosine schedule to $2.5 \times 10^{-5}$. Training proceeds for 240k iterations with a batch size of 1024 rays. Loss weights are set to $\lambda_{\text{eik}}=0.05$, $\lambda_{d}=0.005$, and $\lambda_{n}=0.003$. Monocular depth and normal priors are generated by Omnidata~\cite{Omnidata}.

\begin{figure}
\centering
\includegraphics[width=\columnwidth]{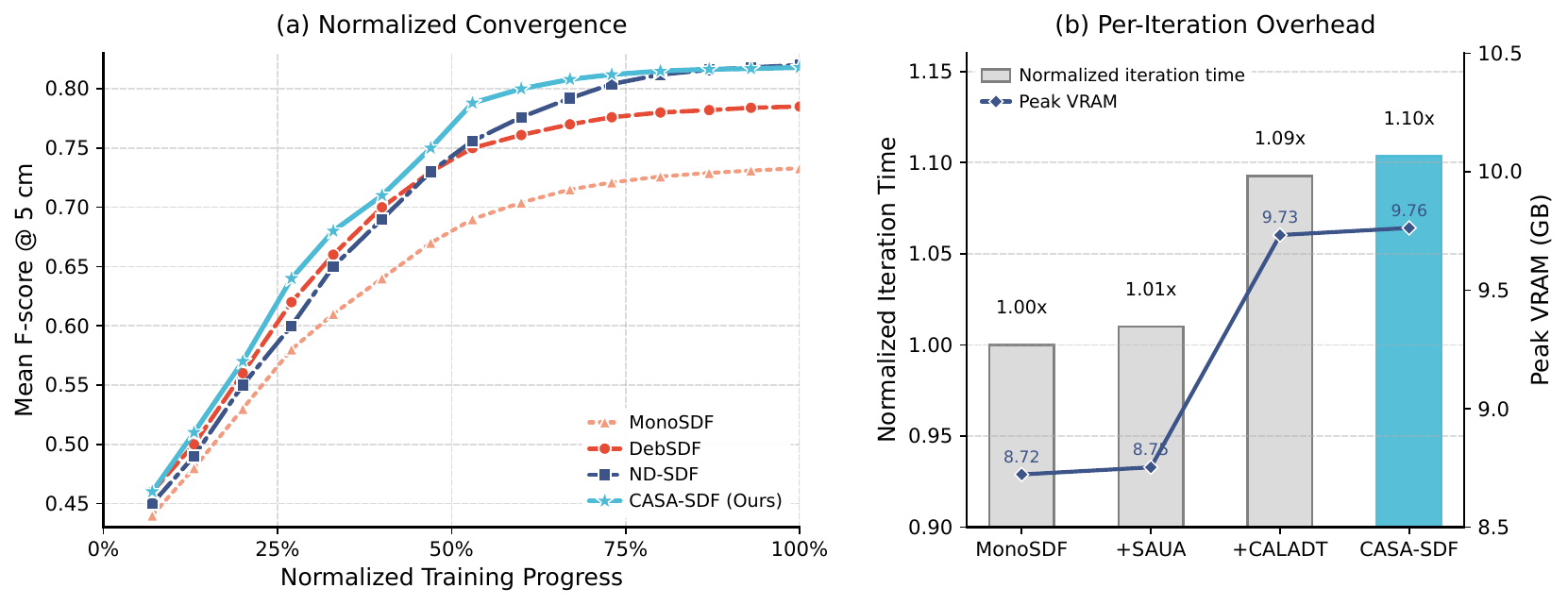}
\caption{Training efficiency and per-iteration overhead analysis on ScanNet under matched optimization budgets. \textbf{(a)} Mean F-score versus optimization iterations. \textbf{(b)} Module-level per-iteration overhead decomposition, where bars denote normalized steady-state iteration time and the line indicates peak GPU memory usage. SAUA uses offline pre-computed uncertainty maps and incurs negligible online overhead, whereas the additional computational cost is primarily attributed to the curvature-aware local density modulation in CALADT.
}
\label{fig:efficiency_profile}
\end{figure}

The training efficiency of CASA-SDF is analyzed in Fig.~\ref{fig:efficiency_profile} 
through two complementary perspectives: convergence behavior and per-iteration 
overhead. Fig.~\ref{fig:efficiency_profile}(a) compares the F-score trajectories 
of the main methods under a matched optimization budget. CASA-SDF exhibits 
stable early-stage optimization, achieves its principal gains during the middle 
stage as adaptive supervision and sharpness modulation take effect, and gradually 
saturates in the later stage.

Fig.~\ref{fig:efficiency_profile}(b) reports the module-level online overhead 
relative to the MonoSDF baseline. SAUA introduces negligible online cost because 
the semantic and photometric uncertainty maps are pre-computed once before 
training and subsequently cached as fixed lookup maps. The monocular depth and 
normal prediction step is shared with prior-based baselines; the CASA-SDF-specific 
pre-computation consists of AngMF confidence normalization and patch-based 
photometric uncertainty estimation. In our implementation, this offline step 
requires several minutes per ScanNet scene, which is small compared with the full 240k-iteration 
training budget and is amortized over the entire optimization process.

The additional online overhead is primarily attributable to CALADT, whose 
curvature-aware modulation requires extra local SDF-gradient evaluations near 
the evolving surface during training. On a single NVIDIA RTX 4090 GPU, the 
baseline runs at 0.226\,s per iteration (4.43\,it/s) with 8.93\,GB peak VRAM, 
whereas the full CASA-SDF pipeline runs at 0.247\,s per iteration (4.05\,it/s) 
with 9.97\,GB peak VRAM. This corresponds to a 9.3\% increase in per-iteration 
runtime and an 11.6\% increase in peak GPU memory. CALADT does not introduce an additional inference network or post-processing stage. Because the final mesh is extracted from the zero-level set of the trained SDF using the same marching-cubes procedure as standard NeuS-style methods, its impact on test-time rendering and mesh extraction is negligible, and the extra computational cost is therefore limited to training-time optimization. Overall, CASA-SDF improves reconstruction quality with a modest increase in computational cost.

\subsection{Comparisons}

\subsubsection{Quantitative Evaluation}

\noindent\textbf{Evaluation on ScanNet.}
Table~\ref{tab:scannet} summarizes the ScanNet results under the standard 5\,cm evaluation protocol.
Since this table aggregates results from disparate publications and codebases, we treat it as a broad benchmark comparison. To account for potential metric variance arising from different mesh extraction protocols or culling scripts, we complement this with controlled reproduced results in the Supplementary Material.

Small differences in mesh extraction, culling, or training recipes may slightly affect absolute values, even when the same dataset split is used.

As shown in Table~\ref{tab:scannet}, CASA-SDF achieves a highly competitive overall F-score of 0.818, substantially outperforming MonoSDF~\cite{MonoSDF} (0.733) and approaching ND-SDF~\cite{ndsdf} (0.820). We note that ND-SDF remains slightly better in global F-score, Accuracy, and Precision on ScanNet. The main advantage of CASA-SDF lies in Completeness (0.035) and Recall (0.810), where it obtains the best results among all compared methods.

\textit{Discussion on metric trade-offs:}
CASA-SDF is particularly effective at recovering complete surfaces, although its Accuracy (0.033) and Precision (0.827) are slightly lower than those of ND-SDF. This behavior reflects a familiar precision-recall trade-off in indoor implicit surface reconstruction. Methods optimized for very high precision often favor globally conservative surfaces, which suppress noise on dominant planes but may under-reconstruct extremely thin structures. By contrast, CASA-SDF shifts the representation toward greater structural completeness: SAUA reduces the long-term influence of unreliable priors, while CALADT allocates higher local bandwidth to geometrically complex regions. This more aggressive recovery strategy improves Recall and Completeness, at the cost of slightly increased geometric variance around some high-frequency boundaries. We therefore interpret the ScanNet results in conjunction with the region-aware analysis in Sec.~\ref{sec:heterogeneity_analysis}, where improvements on thin structures can be assessed separately from the planar majority.

\noindent\textbf{Evaluation on Replica.}
Table~\ref{tab:replica} summarizes the quantitative comparisons on the synthetic Replica dataset. 
To ensure a consistent baseline comparison, we evaluate against representative methods that provide official configurations or established evaluation protocols for this domain. Certain ScanNet baselines are omitted because their published settings are not directly transferable to Replica in a controlled manner.

Consistent with the ScanNet results, CASA-SDF achieves the best Completeness (0.045), Recall (0.776), and overall F-score (0.788) on Replica. The same trade-off is observed in this synthetic setting: ND-SDF yields slightly better Accuracy and Precision, whereas CASA-SDF recovers more complete geometry on thin and delicate structures. This result suggests that correcting prior bias alone is insufficient under geometric heterogeneity; spatially adaptive representation bandwidth is also required to preserve high-frequency indoor geometry without globally increasing noise.

\begin{table}[t]
\centering
\caption{Quantitative comparisons of room-scale surface reconstruction over 8 ScanNet scenes. CASA-SDF achieves the best Completeness and Recall while maintaining a competitive overall F-score.}
\label{tab:scannet}
\resizebox{\linewidth}{!}{%
\begin{tabular}{lccccc}
\toprule
Method & Acc $\downarrow$ & Comp $\downarrow$ & Prec $\uparrow$ & Rec $\uparrow$ & F-score $\uparrow$ \\
\midrule
COLMAP~\cite{colmap} & 0.047 & 0.235 & 0.711 & 0.441 & 0.537 \\
NeuS~\cite{NeuS} & 0.179 & 0.208 & 0.313 & 0.275 & 0.291 \\
\midrule
NeuRIS~\cite{NeuRIS} & 0.051 & 0.048 & 0.720 & 0.674 & 0.696 \\
MonoSDF~\cite{MonoSDF} & 0.035 & 0.048 & 0.799 & 0.681 & 0.733 \\
HelixSurf~\cite{HelixSurf} & 0.038 & 0.044 & 0.786 & 0.727 & 0.755 \\
\midrule
DebSDF~\cite{Debsdf} & 0.036 & 0.040 & 0.807 & 0.765 & 0.785 \\
Fine-Recon~\cite{Fine-Recon} & 0.033 & 0.041 & 0.814 & 0.737 & 0.773 \\
ND-SDF~\cite{ndsdf} & \textbf{0.031} & 0.036 & \textbf{0.840} & 0.803 & \textbf{0.820} \\
\midrule
\textbf{Ours} & 0.033 & \textbf{0.035} & 0.827 & \textbf{0.810} & 0.818 \\
\bottomrule
\end{tabular}%
}
\end{table}

\begin{table}[t]
\centering
\caption{Quantitative comparisons of room-scale surface reconstruction over 8 Replica scenes. CASA-SDF achieves the best Completeness, Recall, and overall F-score. Following DebSDF~\cite{Debsdf}, the implicit-representation baselines are reported under their MLP-only configurations for consistency.}
\label{tab:replica}
\resizebox{\linewidth}{!}{%
\begin{tabular}{lccccc}
\toprule
Method & Acc $\downarrow$ & Comp $\downarrow$ & Prec $\uparrow$ & Rec $\uparrow$ & F-score $\uparrow$ \\
\midrule
NeuRIS~\cite{NeuRIS} & 0.074 & 0.147 & 0.489 & 0.387 & 0.431 \\
MonoSDF~\cite{MonoSDF} & 0.081 & 0.139 & 0.497 & 0.423 & 0.454 \\
occSDF~\cite{occsdf} & 0.052 & 0.075 & 0.715 & 0.642 & 0.676 \\
\midrule
DebSDF~\cite{Debsdf} & 0.044 & 0.051 & 0.786 & 0.725 & 0.754 \\
Fine-Recon~\cite{Fine-Recon} & 0.041 & 0.048 & 0.792 & 0.741 & 0.765 \\
ND-SDF~\cite{ndsdf} & \textbf{0.035} & 0.047 & \textbf{0.812} & 0.758 & 0.784 \\
\midrule
\textbf{Ours} & 0.038 & \textbf{0.045} & 0.801 & \textbf{0.776} & \textbf{0.788} \\
\bottomrule
\end{tabular}%
}
\end{table}

\subsubsection{Qualitative Evaluation}

\begin{figure*}
\centering
\IfFileExists{figures/exp1.pdf}{\includegraphics[width=.95\textwidth]{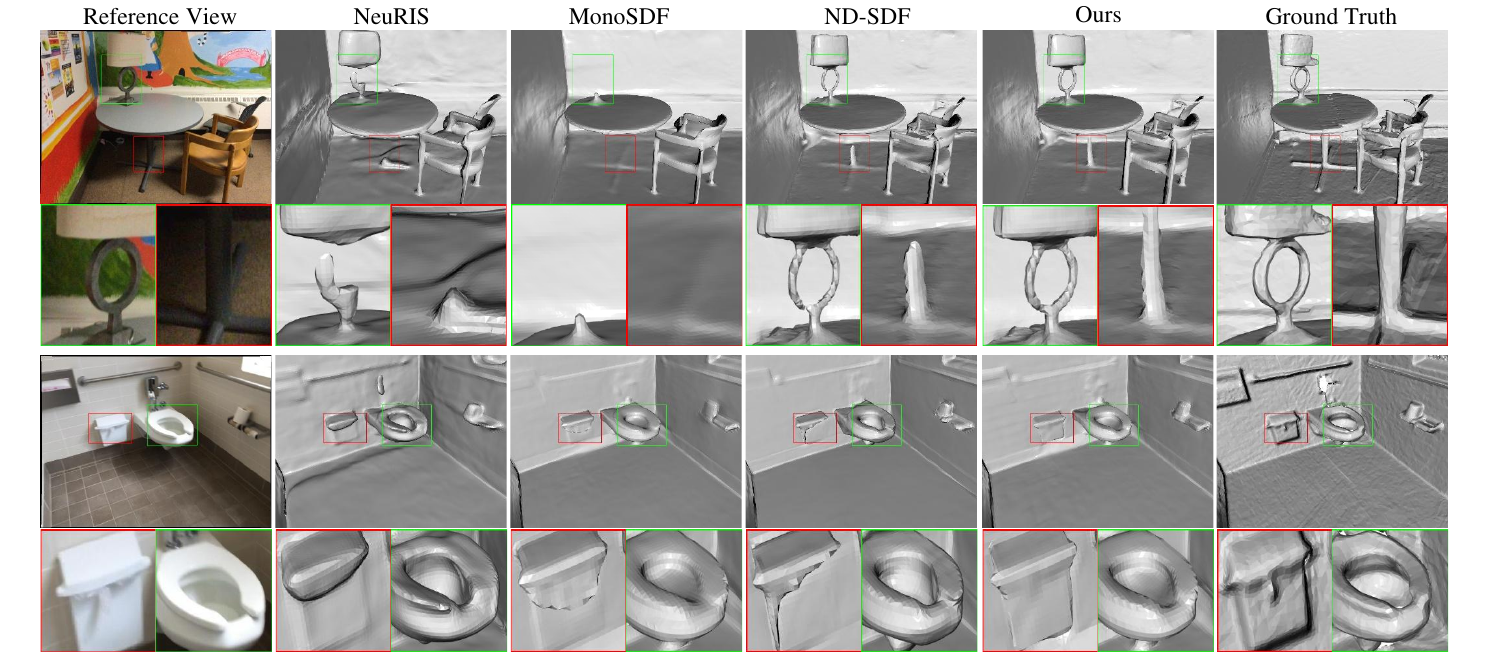}}{}
\caption{Qualitative mesh comparisons on ScanNet~\cite{scanNet} for two representative scenes. CASA-SDF better preserves thin structures (e.g., chair legs), small objects (e.g., lamps), and wall-mounted details, while reducing floating artifacts on texture-less planar regions.}
\label{fig:qualitative_mesh_scannet}
\end{figure*}

\noindent\textbf{ScanNet mesh comparison.}
Fig.~\ref{fig:qualitative_mesh_scannet} visualizes reconstructed meshes for two ScanNet scenes containing thin structures (e.g., chair legs), small indoor objects (e.g., table lamps), and wall-mounted attachments. Compared with MonoSDF and NeuRIS, CASA-SDF better preserves slender geometry while reducing spurious floating fragments on large planar regions.

\begin{figure*}
\centering
\IfFileExists{figures/exp2.pdf}{\includegraphics[width=.95\textwidth]{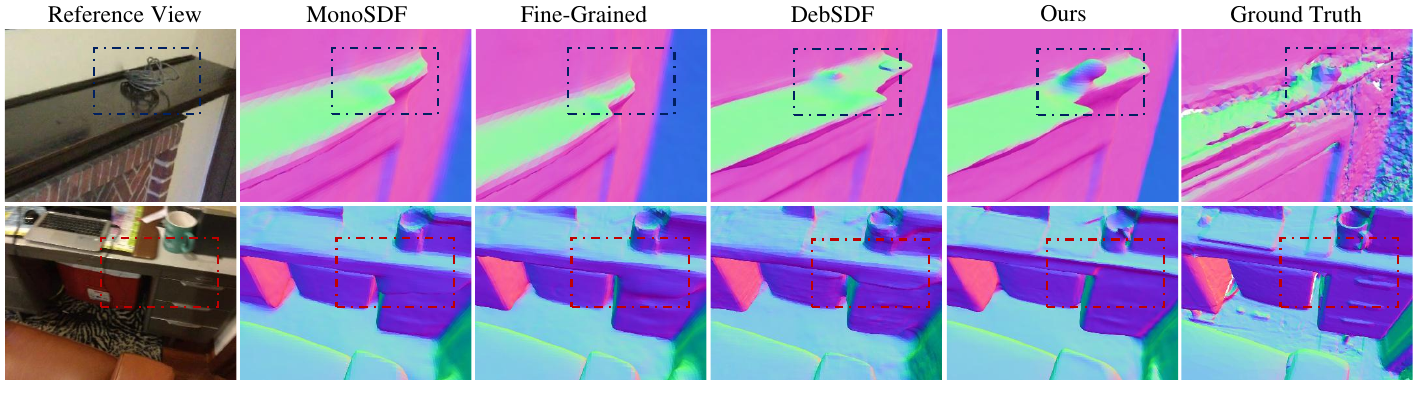}}{}
\caption{Qualitative normal comparisons on ScanNet~\cite{scanNet}. CASA-SDF reconstructs smoother and more coherent normals on tabletop surface regions while preserving sharp boundaries.}
\label{fig:qualitative_normal_scannet}
\end{figure*}

\noindent\textbf{ScanNet normal comparison.}
Fig.~\ref{fig:qualitative_normal_scannet} compares rendered normal maps on a ScanNet scene with challenging table-top details. CASA-SDF reconstructs cleaner and more coherent normal patterns on texture-less surfaces, while maintaining sharp discontinuities around object boundaries.

\begin{figure*}
\centering
\IfFileExists{figures/exp4.pdf}{\includegraphics[width=.95\textwidth]{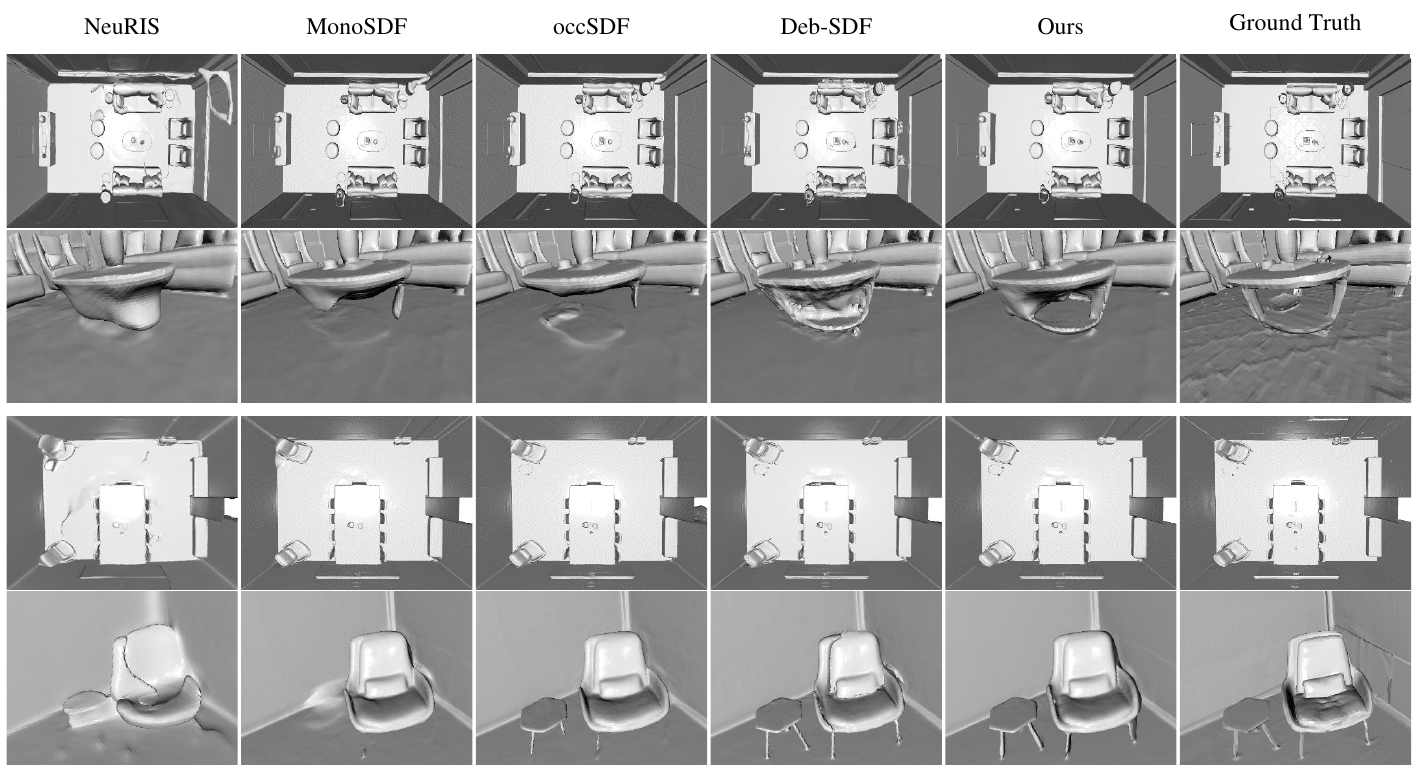}}{}
\caption{Qualitative mesh comparisons on Replica~\cite{Replica}. Top-down views of room-scale geometry and zoomed local reconstructions are shown. Compared with the baselines, CASA-SDF better preserves the global room structure while recovering sharper and more complete fine details on thin furniture parts and wall attachments.}
\label{fig:qualitative_mesh_replica}
\end{figure*}

\begin{figure*}
\centering
\IfFileExists{figures/exp3.pdf}{\includegraphics[width=.95\textwidth]{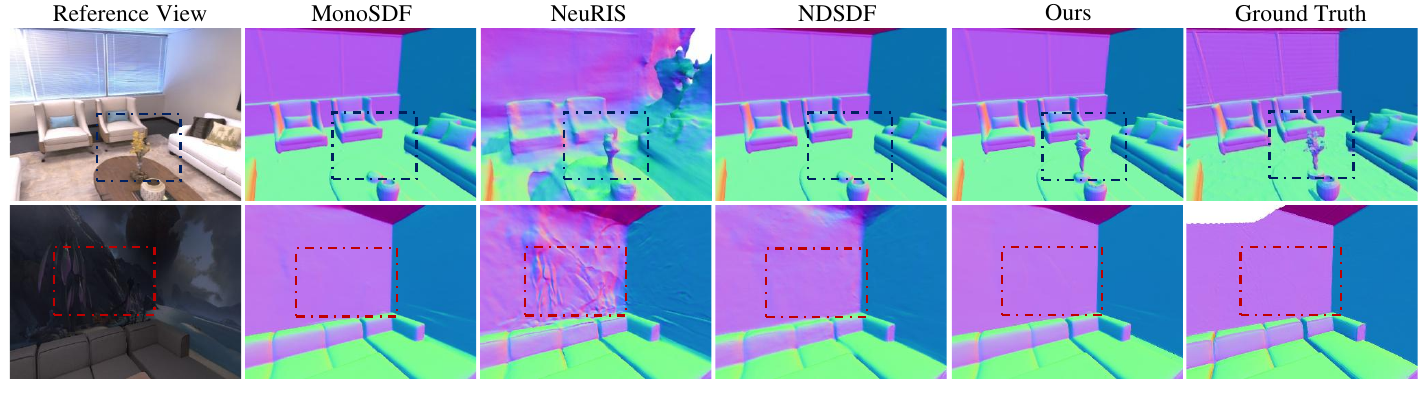}}{}
\caption{Qualitative normal comparisons on Replica~\cite{Replica}. CASA-SDF produces smoother wall and tabletop normals while retaining clear geometric discontinuities.}
\label{fig:qualitative_normal_replica}
\end{figure*}

\noindent\textbf{Replica mesh and normal comparison.}
Fig.~\ref{fig:qualitative_mesh_replica} and Fig.~\ref{fig:qualitative_normal_replica} present qualitative results on Replica. CASA-SDF produces more complete thin structures in the mesh and smoother planar normals (on walls and table tops), demonstrating a better balance between detail preservation and surface regularization.

\subsection{Ablation Studies}
\label{sec:ablation}

We conduct ablation studies on the ScanNet dataset to validate the effectiveness of the proposed mechanisms and the rationale behind their design. Unless otherwise stated, all ablation experiments follow the same training protocol and evaluation script as the main results, ensuring that performance differences can be attributed to the ablated components rather than to protocol variations.

\noindent\textbf{Effectiveness of Core Components.}
We adopt MonoSDF~\cite{MonoSDF} as the baseline, which employs static binary prior filtering and a globally shared sharpness parameter in the SDF-to-density mapping. Table~\ref{tab:ablation_components} incrementally incorporates the two proposed components, with the following observations:
\begin{itemize}
    \item \textbf{+ SAUA}: Replacing the static filtering with Spatially-Adaptive Uncertainty Annealing (SAUA) improves the F-score from 0.733 to 0.785, primarily by enhancing Completeness and Recall under unreliable prior conditions. This confirms that adaptive prior scheduling effectively mitigates the negative impact of noisy monocular cues.
    \item \textbf{+ CALADT}: Applying Curvature-Aware Locally Adaptive Density Transformation (CALADT) yields an F-score of 0.755, accompanied by improved Completeness ($0.048 \to 0.044$). This result demonstrates that spatially varying sharpness helps preserve thin and high-curvature structures while avoiding excessive smoothing on planar regions, addressing the inherent trade-off of global sharpness settings.
    \item \textbf{Full Model}: Combining the two mechanisms achieves the best overall performance (0.818 F-score). While SAUA or CALADT alone slightly degrades Accuracy relative to the baseline due to their more aggressive recovery of fine details, their combination yields the strongest overall result. This indicates that adaptive prior scheduling and adaptive representation bandwidth are complementary rather than redundant, as they jointly address the dual challenges of prior unreliability and geometric heterogeneity.
\end{itemize}

\paragraph{Synergistic behavior of SAUA and CALADT.}
Although adding SAUA or CALADT alone slightly degrades Accuracy relative to the baseline (0.036 and 0.038 vs. 0.035), their combination yields the best Accuracy (0.033). This behavior is consistent with a coupled reduction of large geometric deviations in structurally complex regions rather than a uniform improvement across all surface types.

In isolation, SAUA attenuates unreliable priors in thin-structure and boundary regions, yet without additional local representation bandwidth these released regions may still be reconstructed with fragmented geometry or local geometric instabilities. CALADT alone improves local representation capacity, but if unreliable priors remain active in those regions, the sharpened density mapping may instead propagate prior-induced bias into finer-scale detail. When the two modules are jointly active, SAUA suppresses prior-induced bias in unreliable regions, while CALADT supplies the local bandwidth needed to recover geometrically consistent detail. Consequently, the full model mitigates severe local misalignments that would otherwise contribute disproportionately to the mean Accuracy, while preserving competitive stability on the dominant planar regions. This interpretation is also consistent with the region-aware analysis, where the clearest gains of CASA-SDF are observed on transitional and high-frequency structures rather than on the planar majority.

\begin{table}[h]
\centering
\caption{Ablation of the two core components on ScanNet. SAUA and CALADT provide complementary improvements, and their combination yields the best overall result.}
\label{tab:ablation_components}
\resizebox{0.95\linewidth}{!}{%
\begin{tabular}{cccccc}
\toprule
Baseline (MonoSDF) & SAUA & CALADT & Acc $\downarrow$ & Comp $\downarrow$ & F-score $\uparrow$ \\
\midrule
$\checkmark$ & & & 0.035 & 0.048 & 0.733 \\
$\checkmark$ & $\checkmark$ & & 0.036 & 0.040 & 0.785 \\
$\checkmark$ & & $\checkmark$ & 0.038 & 0.044 & 0.755 \\
$\checkmark$ & $\checkmark$ & $\checkmark$ & \textbf{0.033} & \textbf{0.035} & \textbf{0.818} \\
\bottomrule
\end{tabular}%
}
\end{table}

\begin{figure*}
\centering
\includegraphics[width=.95\textwidth]{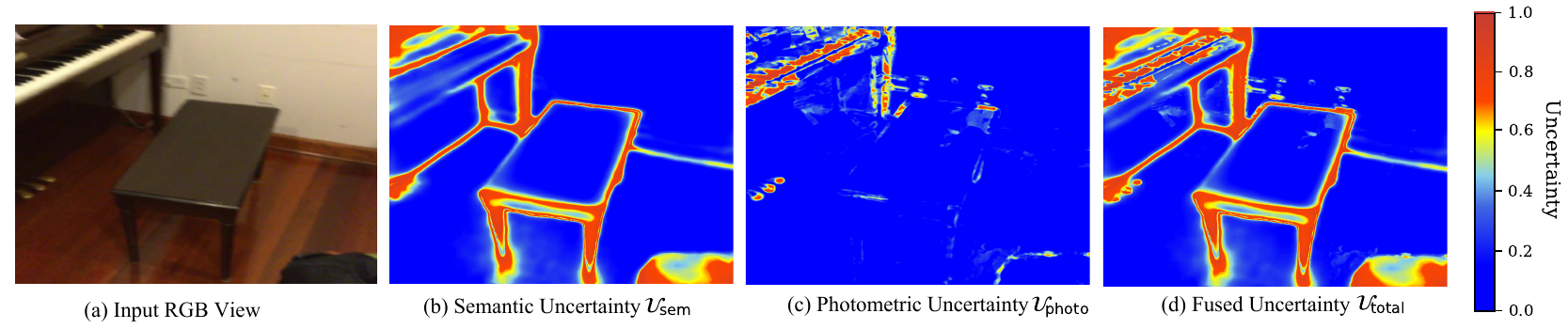}
\caption{Visualization of Hybrid-SAUA uncertainty fusion. (a) Input RGB view. (b) Semantic uncertainty $\mathcal{U}_{\text{sem}}$ highlights boundaries and thin structures where CNN-based priors are unreliable. (c) Photometric uncertainty $\mathcal{U}_{\text{photo}}$ highlights regions where multi-view photometric consistency is weak. (d) The fused uncertainty $\mathcal{U}_{\text{total}}=\max(\mathcal{U}_{\text{sem}},\mathcal{U}_{\text{photo}})$ conservatively marks a pixel as uncertain whenever either source is unreliable, thereby enabling the pixel-wise curriculum in Eq.~\ref{eq:saua}.}
\label{fig:SAUA_Uncertainty}
\end{figure*}

\noindent\textbf{Analysis of Uncertainty Schedules (SAUA).}
A key aspect of SAUA is its time-varying curriculum mechanism. Fig.~\ref{fig:SAUA_Uncertainty} visualizes the semantic and photometric uncertainties along with their conservative fusion strategy. In Table~\ref{tab:ablation_saua}, we compare different weighting strategies for the normal prior loss under the SAUA-only setting (i.e., without CALADT) to isolate the effect of the scheduling mechanism:
\begin{itemize}
    \item \textit{Uniform}: A constant weight $\lambda=1.0$ (NeuRIS style). This leads to high Accuracy (i.e., large reconstruction error) due to over-fitting to noisy priors, as unregulated prior supervision propagates errors from unreliable regions.
    \item \textit{Static Threshold}: Hard binary masking based on uncertainty (MonoSDF style). While this reduces the impact of noisy priors, the rigid binary split fails to adapt to the gradual improvement of geometric estimates during training.
    \item \textit{Inverse Uncertainty}: Static soft weighting using an uncertainty-derived score, but without time decay. This provides a more nuanced prior weighting than binary masking but lacks the adaptive curriculum needed to balance early stabilization and late-stage refinement.
    \item \textit{Annealing (Ours)}: Uncertainty-aware, time-decaying weight as defined in Eq.~\ref{eq:saua}. This dynamic strategy balances early prior-driven stabilization and late-stage relaxation in unreliable regions.
\end{itemize}

The results demonstrate that uncertainty-aware annealing consistently outperforms static schedules. This indicates that prior supervision is most beneficial when it is strong during early training to stabilize coarse geometry, but progressively relaxed in unreliable regions as the neural field accumulates sufficient multi-view evidence for finer refinement. The gradual attenuation prevents the imprinting of biased priors, which becomes increasingly critical once photometric consistency begins to drive data-driven geometric correction.

\begin{table}[h]
\centering
\footnotesize
\setlength{\tabcolsep}{3pt}
\caption{Comparison of prior-weighting schedules under the SAUA-only setting. The proposed time-varying annealing strategy outperforms static alternatives.}
\label{tab:ablation_saua}
\begin{tabular}{lp{3.8cm}cc}
\toprule
Schedule & Description & Acc $\downarrow$ & F-score $\uparrow$ \\
\midrule
Uniform & Constant $w_{\text{prior}}=1$ & 0.045 & 0.710 \\
Static Threshold & Binary mask $w_{\text{prior}}\in\{0,1\}$ & 0.039 & 0.750 \\
Inv-Uncertainty & Static soft weighting (no time decay) & 0.037 & 0.768 \\
\textbf{Annealing (Ours)} & Dynamic uncertainty-aware annealing (Eq.~\ref{eq:saua}) & \textbf{0.036} & \textbf{0.785} \\
\bottomrule
\end{tabular}
\end{table}

\noindent\textbf{Contribution of Uncertainty Sources (SAUA).}
To validate the complementarity of semantic and photometric uncertainties in SAUA, we ablate different uncertainty sources for constructing $\mathcal{U}_{\text{total}}$ \emph{under the SAUA-only setting}. The results in Table~\ref{tab:ablation_uncertainty_sources} clearly demonstrate that fusing both uncertainty sources yields the strongest performance, confirming that semantic and photometric cues provide complementary information about prior reliability.

\begin{table}[h]
\centering
\caption{Ablation of uncertainty sources for SAUA. Using both semantic and photometric uncertainty yields the strongest result.}
\label{tab:ablation_uncertainty_sources}
\begin{tabular}{lcc}
\toprule
Setting & Acc $\downarrow$ & F-score $\uparrow$ \\
\midrule
Semantic only ($\mathcal{U}_{\text{total}}=\mathcal{U}_{\text{sem}}$) & 0.038 & 0.765 \\
Photometric only ($\mathcal{U}_{\text{total}}=\mathcal{U}_{\text{photo}}$) & 0.039 & 0.758 \\
Both ($\mathcal{U}_{\text{total}}=\max(\mathcal{U}_{\text{sem}},\mathcal{U}_{\text{photo}})$) & \textbf{0.036} & \textbf{0.785} \\
\bottomrule
\end{tabular}
\end{table}

\noindent\textbf{Uncertainty versus Curvature for Sharpness Modulation.}
A natural question is whether the semantic uncertainty estimated by AngMF can serve as a direct substitute for the curvature proxy in CALADT. To examine this rigorously, Table~\ref{tab:ablation_signal} compares three modulation signals under identical training conditions, reporting the full metric set. The AngMF-driven variant improves over fixed sharpness, confirming that uncertainty encodes useful information about visually complex regions. Nevertheless, the curvature-driven variant achieves the best overall performance.

The performance gains are not uniform across metrics: relative to AngMF uncertainty, curvature yields higher Completeness, Recall, and F-score, at the cost of slightly reduced Accuracy and Precision. This pattern reflects the distinct roles of the two signals: semantic uncertainty acts primarily as a 2D prior-reliability cue, whereas curvature serves as a 3D geometry-complexity cue derived from the evolving SDF field. Consequently, curvature is better aligned with the local bandwidth requirements of the implicit representation. Specifically, curvature is more effective at allocating representation bandwidth to thin and geometrically challenging structures, even at the cost of slightly increased boundary variance---a trade-off that favors indoor reconstruction.
\begin{table}[h]
\centering
\setlength{\tabcolsep}{3pt}
\caption{Ablation of the signal used to modulate sharpness in CALADT. Among the tested alternatives, curvature provides the strongest overall trade-off, with higher Completeness, Recall, and F-score.}
\label{tab:ablation_signal}
\begin{tabular}{lccccc}
\toprule
Modulation signal & Acc $\downarrow$ & Comp $\downarrow$ & Prec $\uparrow$ & Rec $\uparrow$ & F-score $\uparrow$ \\
\midrule
Fixed sharpness & \textbf{0.035} & 0.048 & \textbf{0.799} & 0.681 & 0.733 \\
AngMF uncertainty & 0.037 & 0.045 & 0.793 & 0.718 & 0.753 \\
\textbf{Curvature (Ours)} & 0.038 & \textbf{0.044} & 0.790 & \textbf{0.732} & \textbf{0.755} \\
\bottomrule
\end{tabular}
\end{table}

\begin{figure*}[!t]
\centering
\includegraphics[width=.95\textwidth]{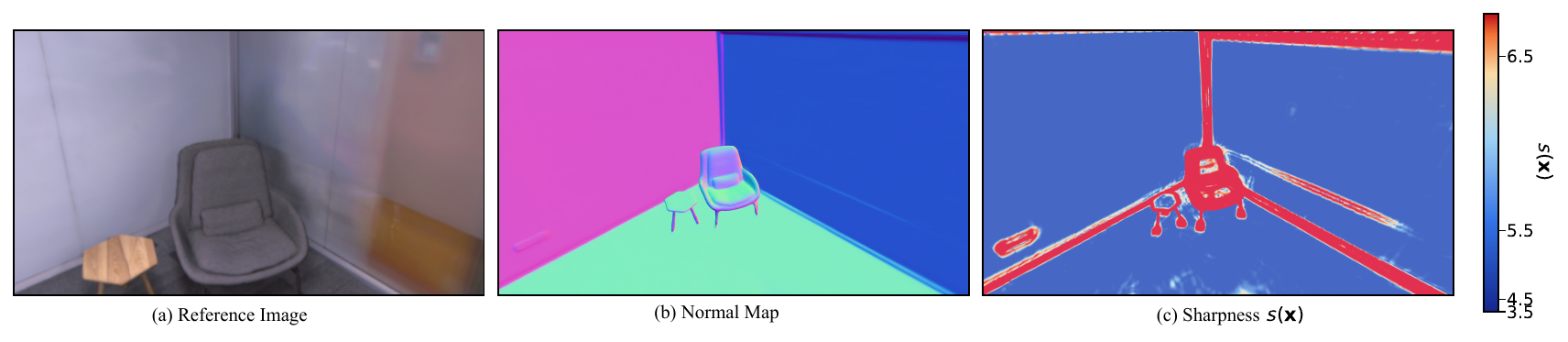}
\caption{Visualization of the spatially adaptive sharpness field $s(\mathbf{x})$. CALADT assigns stronger smoothing (small $s$) to planar regions and stronger sharpening (large $s$) to detailed structures.}
\label{fig:sharpness}
\end{figure*}

\noindent\textbf{Visualization of Adaptive Sharpness.}
To illustrate the behavior of CALADT, we visualize the learned sharpness field $s(\mathbf{x})$ in Fig.~\ref{fig:sharpness}. 
The model assigns small $s$ values (blue) to walls and floors to suppress planar noise, while assigning large $s$ values (red) to chair 
legs and table edges to preserve high-frequency structures. This visualization directly confirms that CALADT adaptively allocates representation bandwidth 
according to local geometric complexity.

\subsection{Analysis of Geometric Heterogeneity}
\label{sec:heterogeneity_analysis}

To explicitly assess the impact of spatial adaptation on diverse geometric structures, we conduct a region-aware decoupled evaluation on the ScanNet 
dataset.
\paragraph{Region-Aware Evaluation Protocol.}
We define geometric partitions exclusively on the ground-truth (GT) mesh to ensure all methods are evaluated against the same reference, independent of predicted topology or local noise. Specifically, we estimate a normalized mean-curvature magnitude on the GT surface using a cotangent-Laplacian operator, and assign each triangle a curvature score by averaging values at its incident vertices. To reduce tessellation bias, statistics are computed over area-weighted triangles rather than raw vertex counts.

To establish robust partition criteria, we pool area-weighted GT triangles from all 8 ScanNet evaluation scenes and compute percentile thresholds once on the pooled distribution. These fixed thresholds are applied uniformly: \textit{Planar Regions} correspond to the lowest-curvature 75\%, \textit{High-Frequency Structures} to the highest-curvature 10\%, and the remaining 15\% form \textit{Transitional Regions}. This reflects the long-tail distribution typical of indoor scenes, where large planar surfaces dominate area while thin parts and sharp boundaries occupy a small fraction. Sensitivity to nearby alternative thresholds is reported in the Supplementary Material.

Regional metrics are computed as follows. Completeness and Recall---defined on the ground-truth surface---are evaluated directly within each partition. For Accuracy and Precision, each predicted point inherits the region label of its nearest GT triangle, ensuring GT-to-prediction and prediction-to-GT measurements share the same spatial reference. Table~\ref{tab:region_metrics} reports the resulting regional F-scores. CASA-SDF achieves its clearest gains in transitional and high-frequency regions, where geometric complexity is concentrated. On planar regions, ND-SDF attains the highest score, while CASA-SDF remains competitive. This indicates that adaptive sharpness enhances recovery of geometrically challenging structures without compromising planar stability. Under geometric heterogeneity, correcting prior bias alone is insufficient; explicit spatial adaptation of representation bandwidth provides additional benefits for recovering delicate thin structures. Accordingly, the global Accuracy improvement observed in the full model (Table~\ref{tab:ablation_components}) is driven by the reduction of large outlier distances in high-frequency regions, rather than by a global precision advantage on planar surfaces.
\begin{table}[h]
\centering
\caption{Region-aware F-score breakdown on ScanNet under the 5\,cm threshold. All entries correspond to F-score computed within GT-curvature-based partitions. Planar regions dominate the global metric, while CASA-SDF shows the largest relative gain on transitional and high-frequency structures.}
\label{tab:region_metrics}
\resizebox{\linewidth}{!}{%
\begin{tabular}{lcccc}
\toprule
Method & Global F-score & Planar F-score & Transitional F-score & High-Freq F-score \\
 & (100\%) & ($\sim$75\%) & ($\sim$15\%) & ($\sim$10\%) \\
\midrule
MonoSDF~\cite{MonoSDF} & 0.733 & 0.812 & 0.676 & 0.587 \\
DebSDF~\cite{Debsdf} & 0.785 & 0.826 & 0.723 & 0.651 \\
ND-SDF~\cite{ndsdf} & \textbf{0.820} & \textbf{0.845} & 0.748 & 0.724 \\
\textbf{Ours} & 0.818 & 0.836 & \textbf{0.759} & \textbf{0.772} \\
\bottomrule
\end{tabular}%
}
\end{table}

\begin{figure*}
\centering
\IfFileExists{figures/analysis_heterogeneity.png}{\includegraphics[width=.95\textwidth]{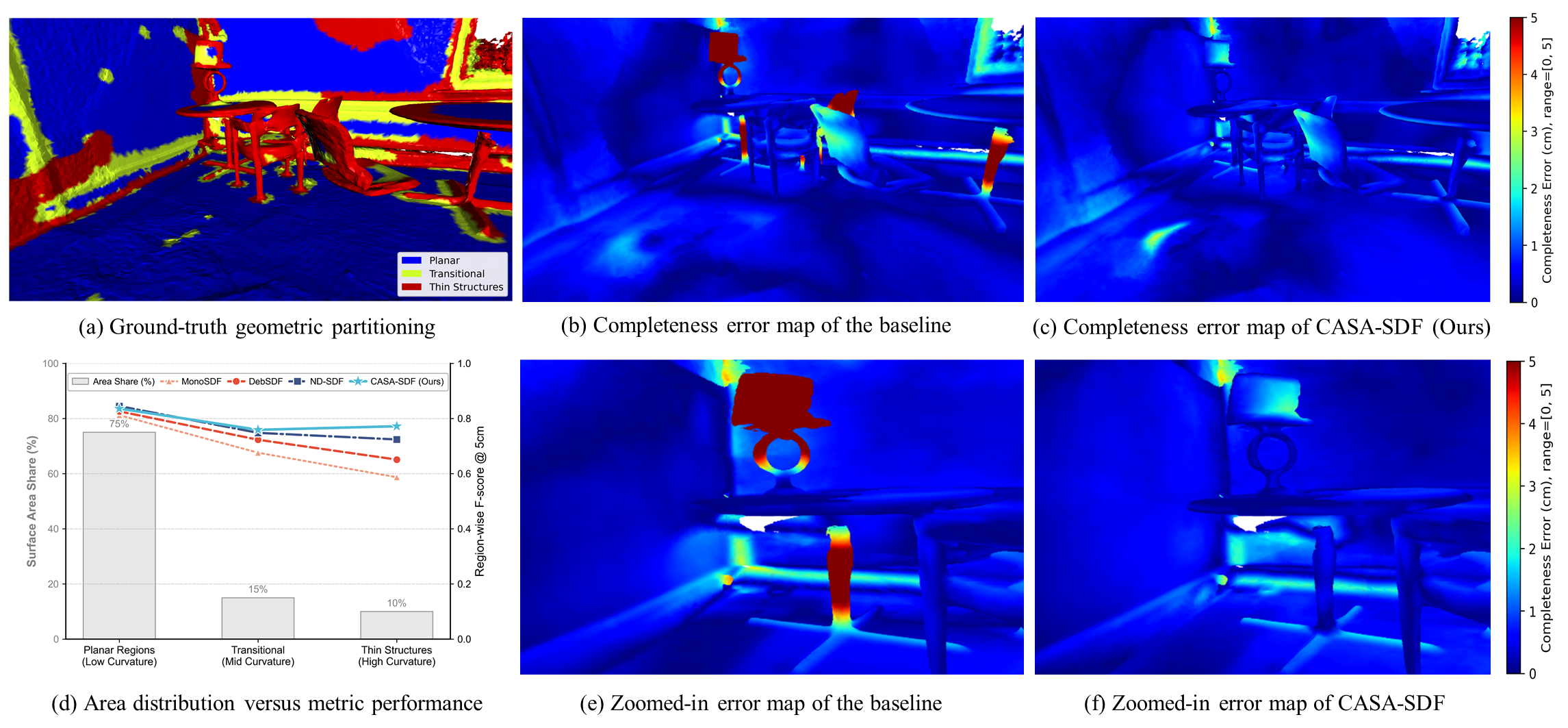}}{}
\caption{Region-aware decoupled evaluation.
\textbf{(a)} Ground-truth geometric partitioning into planar, transitional, and high-frequency regions using the fixed GT-based curvature protocol described in Sec.~\ref{sec:heterogeneity_analysis}.
\textbf{(b)} Completeness error map of a representative baseline (MonoSDF), exhibiting severe truncation errors (red regions, $\ge 5$cm) on thin structures under a globally shared sharpness setting.
\textbf{(c)} Completeness error map of CASA-SDF. The proposed method substantially reduces errors on delicate structures (blue regions) while maintaining planar smoothness.
\textbf{(d)} Area distribution vs. region-aware performance for MonoSDF, DebSDF, ND-SDF, and CASA-SDF, showing that planar regions dominate the scene area, whereas the clearest performance differences emerge in transitional and high-frequency regions.
\textbf{(e)} Zoomed-in MonoSDF error map, revealing over-smoothing and topological breakages on furniture legs.
\textbf{(f)} Zoomed-in CASA-SDF error map, demonstrating improved structural preservation on the same thin components.}
\label{fig:analysis_heterogeneity}
\end{figure*}

\paragraph{Visualizing Decoupled Errors.}
Fig.~\ref{fig:analysis_heterogeneity} visualizes the completeness error ($d_{\text{GT} \to \text{Pred}}$) projected directly onto the GT mesh. To align the visual evidence with the quantitative metric, the colormap is capped at 5\,cm (red), exactly matching the F-score threshold. 
Fig.~\ref{fig:analysis_heterogeneity}(b,e) visualizes MonoSDF as a representative 
baseline exhibiting characteristic failure modes under global sharpness: 
over-smoothing and topological breakages on thin structures. DebSDF and ND-SDF 
are evaluated quantitatively in Table~\ref{tab:region_metrics} and Fig.~\ref{fig:analysis_heterogeneity}(d). Fig.~\ref{fig:analysis_heterogeneity}(c,f) shows that CASA-SDF reduces these localized errors while maintaining smooth planar reconstruction. 

\section{Conclusion}
\label{sec:conclusion}

This paper presents CASA-SDF, a framework that addresses indoor geometric heterogeneity through joint adaptation of supervision lifecycle and representation bandwidth. Evaluations on ScanNet and Replica confirm that this joint adaptation shifts the reconstruction balance toward greater completeness, preserving delicate topological features without introducing instability in large planar regions.
\section*{Declaration of generative AI and AI-assisted technologies in the manuscript preparation process}
During the preparation and revision of this work, the authors used an AI-assisted language and editing tool (such as ChatGPT) to help with language polishing and consistency checking of the manuscript, including alignment with the journal's author guidelines.

\section*{Declaration of competing interest}
The authors declare that they have no known competing financial interests or personal relationships that could have appeared to influence the work reported in this paper.

\section*{Data availability}
Data will be made available on request.

\printcredits
\section*{Acknowledgements}
The authors acknowledge the creators of the ScanNet and Replica datasets, 
and the developers of the Omnidata and AngMF models, for making their 
resources publicly available.
\bibliographystyle{unsrtnat}

\bibliography{ref}

@String(CVPR  = {Proceedings of the IEEE/CVF Conference on Computer Vision and Pattern Recognition (CVPR)})

@String(ICCV  = {Proceedings of the IEEE/CVF International Conference on Computer Vision (ICCV)})

@String(ECCV  = {Proceedings of the European Conference on Computer Vision (ECCV)})

@String(NeurIPS  = {Proceedings of the Advances in Neural Information Processing Systems (NeurIPS)})

@String(ICLR  = {Proceedings of the The International Conference on Learning Representations (ICLR)})

@String(AAAI  = {Proceedings of the AAAI Conference on Artificial Intelligence (AAAI)})

@string{SIGGRAPH  = {Proceedings of the Annual Conference on Computer Graphics and Interactive Techniques (SIGGRAPH)}}

@String(CVPR  = {CVPR})

@String(ICCV  = {ICCV})

@String(ECCV  = {ECCV})

@String(NeurIPS  = {NeurIPS})

@String(ICLR  = {ICLR})

@String(AAAI  = {AAAI})

@string(SIGGRAPH  = {SIGGRAPH})

@article{Nerf,
  title={Nerf: Representing scenes as neural radiance fields for view synthesis},
  author={Mildenhall, Ben and Srinivasan, Pratul P and Tancik, Matthew and Barron, Jonathan T and Ramamoorthi, Ravi and Ng, Ren},
  journal={Communications of the ACM},
  volume={65},
  number={1},
  pages={99--106},
  year={2021},
  publisher={ACM New York, NY, USA}
}

@inproceedings{HelixSurf,
  title={Helixsurf: A robust and efficient neural implicit surface learning of indoor scenes with iterative intertwined regularization},
  author={Liang, Zhihao and Huang, Zhangjin and Ding, Changxing and Jia, Kui},
  booktitle={Proceedings of the IEEE/CVF Conference on Computer Vision and Pattern Recognition},
  pages={13165--13174},
  year={2023}
}

@inproceedings{NeuRIS,
  title={Neuris: Neural reconstruction of indoor scenes using normal priors},
  author={Wang, Jiepeng and Wang, Peng and Long, Xiaoxiao and Theobalt, Christian and Komura, Taku and Liu, Lingjie and Wang, Wenping},
  booktitle={European Conference on Computer Vision},
  pages={139--155},
  year={2022},
  organization={Springer}
}

@article{MonoSDF,
  title={Monosdf: Exploring monocular geometric cues for neural implicit surface reconstruction},
  author={Yu, Zehao and Peng, Songyou and Niemeyer, Michael and Sattler, Torsten and Geiger, Andreas},
  journal={Advances in neural information processing systems},
  volume={35},
  pages={25018--25032},
  year={2022}
}

@article{NeuS,  
 title={NeuS: Learning Neural Implicit Surfaces by Volume Rendering for Multi-view Reconstruction}, 
 journal={arXiv: Computer Vision and Pattern Recognition,arXiv: Computer Vision and Pattern Recognition}, 
 author={Wang, Peng and Liu, Lingjie and Liu, Yuan and Theobalt, Christian and Komura, Taku and Wang, Wenping}, 
 year={2021}, 
 month={Jun}, 
 language={en-US} 
 }

@article{volsdf,
  title={Volume rendering of neural implicit surfaces},
  author={Yariv, Lior and Gu, Jiatao and Kasten, Yoni and Lipman, Yaron},
  journal={Advances in Neural Information Processing Systems},
  volume={34},
  pages={4805--4815},
  year={2021}
}

@inproceedings{HF-neus,
title={{HF}-NeuS: Improved Surface Reconstruction Using High-Frequency Details},
author={Yiqun Wang and Ivan Skorokhodov and Peter Wonka},
booktitle={Advances in Neural Information Processing Systems},
editor={Alice H. Oh and Alekh Agarwal and Danielle Belgrave and Kyunghyun Cho},
year={2022},
url={https://openreview.net/forum?id=UPnJuDKqOfX}
}

@InProceedings{Manhattansdf,
    author    = {Guo, Haoyu and Peng, Sida and Lin, Haotong and Wang, Qianqian and Zhang, Guofeng and Bao, Hujun and Zhou, Xiaowei},
    title     = {Neural 3D Scene Reconstruction With the Manhattan-World Assumption},
    booktitle = {Proceedings of the IEEE/CVF Conference on Computer Vision and Pattern Recognition (CVPR)},
    month     = {June},
    year      = {2022},
    pages     = {5511-5520}
}

@article{Depth,  
 title={Depth Map Prediction from a Single Image using a Multi-Scale Deep Network}, 
 journal={Cornell University - arXiv,Cornell University - arXiv}, 
 author={Eigen, David and Puhrsch, Christian and Fergus, Rob}, 
 year={2014}, 
 month={Jun}, 
 language={en-US} 
 }

@inproceedings{Omnidata,  
 title={Omnidata: A Scalable Pipeline for Making Multi-Task Mid-Level Vision Datasets from 3D Scans}, 
 url={http://dx.doi.org/10.1109/iccv48922.2021.01061}, 
 DOI={10.1109/iccv48922.2021.01061}, 
 booktitle={2021 IEEE/CVF International Conference on Computer Vision (ICCV)}, 
 author={Eftekhar, Ainaz and Sax, Alexander and Malik, Jitendra and Zamir, Amir}, 
 year={2021}, 
 month={Oct}, 
 language={en-US} 
 }

@inproceedings{EEAU,  
 title={Estimating and Exploiting the Aleatoric Uncertainty in Surface Normal Estimation}, 
 url={http://dx.doi.org/10.1109/iccv48922.2021.01289}, 
 DOI={10.1109/iccv48922.2021.01289}, 
 booktitle={2021 IEEE/CVF International Conference on Computer Vision (ICCV)}, 
 author={Bae, Gwangbin and Budvytis, Ignas and Cipolla, Roberto}, 
 year={2021}, 
 month={Oct}, 
 language={en-US} 
 }

@inproceedings{scanNet,  
 title={ScanNet: Richly-annotated 3D Reconstructions of Indoor Scenes}, 
 url={http://dx.doi.org/10.1109/cvpr.2017.261}, 
 DOI={10.1109/cvpr.2017.261}, 
 booktitle={2017 IEEE Conference on Computer Vision and Pattern Recognition (CVPR)}, 
 author={Dai, Angela and Chang, Angel X. and Savva, Manolis and Halber, Maciej and Funkhouser, Thomas and Niessner, Matthias}, 
 year={2017}, 
 month={Jul}, 
 language={en-US} 
 }

@inproceedings{colmap,
  title={Pixelwise view selection for unstructured multi-view stereo},
  author={Sch{\"o}nberger, Johannes L and Zheng, Enliang and Frahm, Jan-Michael and Pollefeys, Marc},
  booktitle={Computer Vision--ECCV 2016: 14th European Conference, Amsterdam, The Netherlands, October 11-14, 2016, Proceedings, Part III 14},
  pages={501--518},
  year={2016},
  organization={Springer}
}

@article{Debsdf,
  title={Debsdf: Delving into the details and bias of neural indoor scene reconstruction},
  author={Xiao, Yuting and Xu, Jingwei and Yu, Zehao and Gao, Shenghua},
  journal={IEEE Transactions on Pattern Analysis and Machine Intelligence},
  year={2024},
  publisher={IEEE}
}

@article{Replica,  
 title={The Replica Dataset: A Digital Replica of Indoor Spaces.}, 
 journal={Cornell University - arXiv,Cornell University - arXiv}, 
 author={Straub, Julian and Whelan, ThomasJ. and Ma, Lingni and Chen, Yufan and Wijmans, Erik and Green, Simon and Engel, Jakob and Mur-Artal, Raul and Ren, CarlYuheng and Verma, Sandeep and Clarkson, Anton and Yan, Mingfei and Budge, Brian and Yan, Ying and Pan, Xiaqing and Yon, June and Zou, Yuyang and Leon, Kimberly and Carter, Nigel and Briales, Jesus and Gillingham, Tyler and Mueggler, Elias and Pesqueira, Luis and Savva, Manolis and Batra, Dhruv and Strasdat, Hauke and Nardi, RenzoDe and Goesele, Michael and Lovegrove, Steven and Newcombe, Richard}, 
 year={2019}, 
 month={Jun}, 
 language={en-US} 
 }

@ARTICLE{IPP,
  author={Zhou, Xiaowei and Guo, Haoyu and Peng, Sida and Xiao, Yuxi and Lin, Haotong and Wang, Qianqian and Zhang, Guofeng and Bao, Hujun},
  journal={IEEE Transactions on Pattern Analysis and Machine Intelligence}, 
  title={Neural 3D Scene Reconstruction With Indoor Planar Priors}, 
  year={2024},
  volume={46},
  number={9},
  pages={6355-6366},
  doi={10.1109/TPAMI.2024.3379833}}

@inproceedings{PMVC,
  title={PMVC: Promoting Multi-View Consistency for 3D Scene Reconstruction},
  author={Zhang, Chushan and Tong, Jinguang and Lin, Tao Jun and Nguyen, Chuong and Li, Hongdong},
  booktitle={Proceedings of the IEEE/CVF Winter Conference on Applications of Computer Vision},
  pages={3678--3688},
  year={2024}
}

@InProceedings{Sun_2024,
    author    = {Sun, Xiaotian and Xu, Qingshan and Yang, Xinjie and Zang, Yu and Wang, Cheng},
    title     = {Global and Hierarchical Geometry Consistency Priors for Few-shot NeRFs in Indoor Scenes},
    booktitle = {Proceedings of the IEEE/CVF Conference on Computer Vision and Pattern Recognition (CVPR)},
    month     = {June},
    year      = {2024},
    pages     = {20530-20539}
}

@ARTICLE{PSDF,
  author={Su, Wanjuan and Zhang, Chen and Xu, Qingshan and Tao, Wenbing},
  journal={IEEE Transactions on Visualization and Computer Graphics}, 
  title={PSDF: Prior-Driven Neural Implicit Surface Learning for Multi-view Reconstruction}, 
  year={2024},
  volume={},
  number={},
  pages={1-16},
  doi={10.1109/TVCG.2024.3444035}}

@inproceedings{NC-SDF,
  title={NC-SDF: Enhancing Indoor Scene Reconstruction Using Neural SDFs with View-Dependent Normal Compensation},
  author={Chen, Ziyi and Wu, Xiaolong and Zhang, Yu},
  booktitle={Proceedings of the IEEE/CVF Conference on Computer Vision and Pattern Recognition},
  pages={5155--5165},
  year={2024}
}

@article{Fine-Recon,
author={Ye, Sheng and Hu, Yubin and Lin, Matthieu and Wen, Yu-Hui and Zhao, Wang and Liu, Yong-Jin and Wang, Wenping},
journal={ IEEE Transactions on Visualization \& Computer Graphics },
title={{ Indoor Scene Reconstruction With Fine-Grained Details Using Hybrid Representation and Normal Prior Enhancement }},
year={2025},
volume={31},
number={09},
ISSN={1941-0506},
pages={5275-5287}}

@article{RayDistance,
  title={RayDistance Volume Rendering for Neural Scene Reconstruction},
  author={Yin, Ruihong and Chen, Yunlu and Karaoglu, Sezer and Gevers, Theo},
  journal={arXiv preprint arXiv:2408.15524},
  year={2024}
}

@inproceedings{Differentiable,
  title={Differentiable volumetric rendering: Learning implicit 3d representations without 3d supervision},
  author={Niemeyer, Michael and Mescheder, Lars and Oechsle, Michael and Geiger, Andreas},
  booktitle={Proceedings of the IEEE/CVF conference on computer vision and pattern recognition},
  pages={3504--3515},
  year={2020}
}

@inproceedings{nerfprior,
  title={NeRFPrior: Learning neural radiance field as a prior for indoor scene reconstruction},
  author={Zhang, Wenyuan and Jia, Emily Yue-ting and Zhou, Junsheng and Ma, Baorui and Shi, Kanle and Liu, Yu-Shen and Han, Zhizhong},
  booktitle={Proceedings of the Computer Vision and Pattern Recognition Conference},
  pages={11317--11327},
  year={2025}
}

@inproceedings{
ndsdf,
title={{ND}-{SDF}: Learning Normal Deflection Fields for High-Fidelity Indoor Reconstruction},
author={Ziyu Tang and Weicai Ye and Yifan Wang and Di Huang and Hujun Bao and Tong He and Guofeng Zhang},
booktitle={The Thirteenth International Conference on Learning Representations},
year={2025},
url={https://openreview.net/forum?id=4HRRcqE9SU}
}

@inproceedings{occsdf,
title={Learning A Room with the Occ-SDF Hybrid: Signed Distance Function Mingled with Occupancy Aids Scene Representation},
author={Xiaoyang Lyu and Peng Dai and Zizhang Li and Dongyu Yan and Yi Lin and Yifan Peng and Xiaojuan Qi},
booktitle={Proceedings of the IEEE/CVF international conference on computer vision},
year={2023}
}

@inproceedings{neuralangelo,
  title={Neuralangelo: High-Fidelity Neural Surface Reconstruction},
  author={Li, Zhaoshuo and M\"uller, Thomas and Evans, Alex and Taylor, Russell H and Unberath, Mathias and Liu, Ming-Yu and Lin, Chen-Hsuan},
  booktitle={IEEE Conference on Computer Vision and Pattern Recognition ({CVPR})},
  year={2023}
}

@article{Sat-DN,
    title={Sat-DN: Implicit Surface Reconstruction from Multi-View Satellite Images with Depth and Normal Supervision},
    author={Liu, Tianle and Zhao, Shuangming and Jiang, Wanshou and Guo, Bingxuan},
    journal={arXiv preprint arXiv:2502.08352},
    year={2025}
}

@misc{GeoPriorWild,
      title={Geometric Prior-Guided Neural Implicit Surface Reconstruction in the Wild}, 
      author={Lintao Xiang and Hongpei Zheng and Bailin Deng and Hujun Yin},
      year={2025},
      eprint={2505.07373},
      archivePrefix={arXiv},
      primaryClass={cs.CV},
      url={https://arxiv.org/abs/2505.07373}, 
}

@misc{BayesSDF,
      title={BayesSDF: Surface-Based Laplacian Uncertainty Estimation for 3D Geometry with Neural Signed Distance Fields}, 
      author={Rushil Desai},
      year={2025},
      eprint={2507.06269},
      archivePrefix={arXiv},
      primaryClass={cs.CV},
      url={https://arxiv.org/abs/2507.06269}, 
}

@inproceedings{IDR_2020,
  title={Multiview Neural Surface Reconstruction by Disentangling Geometry and Appearance},
  author={Yariv, Lior and Kasten, Yoni and others},
  booktitle={NeurIPS},
  year={2020}
}

@inproceedings{NeuralWarp,
  title={Improving Neural Implicit Surfaces Geometry with Patch Warping},
  author={Darmon, Fran{\c{c}}ois and Bascle, B{\'e}n{\'e}dicte and others},
  booktitle={CVPR},
  year={2022}
}

@inproceedings{Ref-NeuS,
  title={Ref-NeuS: Ambiguity-Reduced Neural Implicit Surface Learning for Indoor Scenes with Reflection},
  author={Ge, Wenhang and Hu, Tao and others},
  booktitle={ICCV},
  year={2023}
}

@inproceedings{marigold,
  title={Repurposing Diffusion-Based Image Generators for Monocular Depth Estimation},
  author={Ke, Bingxin and others},
  booktitle={CVPR},
  year={2024}
}

@article{depthanythingv2,
  title={Depth Anything V2},
  author={Yang, Lihe and others},
  journal={arXiv preprint arXiv:2406.09414},
  year={2024}
}

@inproceedings{ActiveNeRF,
  title={ActiveNeRF: Learning where to See with Uncertainty Estimation},
  author={Pan, Xuran and others},
  booktitle={ECCV},
  year={2022}
}

@inproceedings{FisherRF,
  title={FisherRF: Active View Selection via Fisher Information in Neural Radiance Fields},
  author={Jiang, Yuang and others},
  booktitle={CVPR},
  year={2024}
}

@inproceedings{CF-NeRF,
  title={CF-NeRF: Camera Parameter Free Neural Radiance Fields with Probabilistic Uncertainty},
  author={Shen, Yuxin and others},
  booktitle={ECCV},
  year={2022}
}

@inproceedings{ActiveRMAP,
  title={ActiveRMAP: Radiance Field for Active Mapping And Planning},
  author={Gao, Tianruo and others},
  booktitle={ICRA},
  year={2023}
}

@article{StochasticNeRF,
  title={Stochastic Neural Radiance Fields: Quantifying Uncertainty in Implicit 3D Representations},
  author={Jianxiong Shen and Adria Ruiz and Antonio Agudo and Francesc Moreno-Noguer},
  journal={2021 International Conference on 3D Vision (3DV)},
  year={2021},
  pages={972-981},
  url={https://api.semanticscholar.org/CorpusID:237420366}
}

@inproceedings{NeuSample,
  title={NeuSample: Neural Sample Importance for Neural Radiance Fields},
  author={Fang, Jiemin and others},
  booktitle={CVPR},
  year={2024}
}

@inproceedings{StreetSurf,
  title={StreetSurf: Extending Multi-view Implicit Surface Reconstruction to Street Views},
  author={Guo, Haoyu and others},
  booktitle={ICLR},
  year={2024}
}

@inproceedings{S3IM,
  title={S3IM: Stochastic Structural SIMilarity and Its Unreasonable Effectiveness for Neural Fields},
  author={Xie, Zehao and others},
  booktitle={ICCV},
  year={2023}
}

@inproceedings{RegSDF,
  title={RegSDF: Regularizing Neural Surface Reconstruction with Surface-based Geometric Consistency},
  author={Zhang, Jingyang and others},
  booktitle={CVPR},
  year={2024}
}

@inproceedings{Voxurf,
  title={Voxurf: Voxel-based Efficient and Accurate Neural Surface Reconstruction},
  author={Wu, Tong and others},
  booktitle={ICLR},
  year={2023}
}

@inproceedings{PermutoSDF,
  title={PermutoSDF: Fast Multi-View Reconstruction with Implicit Surfaces using Permutohedral Lattices},
  author={Rosu, Radu and others},
  booktitle={CVPR},
  year={2023}
}

@inproceedings{BakedSDF,
  title={BakedSDF: Meshing Neural SDFs for Real-Time View Synthesis},
  author={Yariv, Lior and others},
  booktitle={SIGGRAPH},
  year={2023}
}

@inproceedings{WildFusion,
  title={WildFusion: Learning 3D-Aware Latent Diffusion for View Synthesis in the Wild},
  author={Kolkin, Nicholas and others},
  booktitle={CVPR},
  year={2024}
}

@inproceedings{Wonder3D,
  title={Wonder3D: Single Image to 3D using Cross-Domain Diffusion},
  author={Long, Xiaoxiao and others},
  booktitle={CVPR},
  year={2024}
}

@inproceedings{PixelNeRF,
  title={PixelNeRF: Neural Radiance Fields from One or Few Images},
  author={Yu, Alex and others},
  booktitle={CVPR},
  year={2021}
}

@inproceedings{MVSNeRF,
  title={MVSNeRF: Fast Generalizable Radiance Field Reconstruction from Multi-View Stereo},
  author={Chen, Anpei and others},
  booktitle={ICCV},
  year={2021}
}

@inproceedings{SparseNeuS,
  title={SparseNeuS: Fast Generalizable Neural Surface Reconstruction from Sparse Views},
  author={Long, Xiaoxiao and others},
  booktitle={ECCV},
  year={2022}
}

@inproceedings{Sparis,
    title={Sparis: Neural Implicit Surface Reconstruction of Indoor Scenes from Sparse Views},
    author={Yulun Wu and Han Huang and Wenyuan Zhang and Chao Deng and Ge Gao and Ming Gu and Yu-Shen Liu},
    booktitle={Proceedings of the AAAI Conference on Artificial Intelligence},
    year={2025}
}

@inbook{Pais_2024,
   title={A Probability-Guided Sampler for Neural Implicit Surface Rendering},
   ISBN={9783031729133},
   ISSN={1611-3349},
   url={http://dx.doi.org/10.1007/978-3-031-72913-3_10},
   DOI={10.1007/978-3-031-72913-3_10},
   booktitle={Computer Vision – ECCV 2024},
   publisher={Springer Nature Switzerland},
   author={Pais, Gonçalo Dias and Piedade, Valter and Chatterjee, Moitreya and Greiff, Marcus and Miraldo, Pedro},
   year={2024},
   month=dec, pages={164–182} }

\end{document}